\documentclass[runningheads]{llncs}

\usepackage{eccv}

\makeatletter
\def\@fnsymbol#1{\ensuremath{\ifcase#1\or \dagger\or *\or
   \mathsection\or \mathparagraph\or \|\or **\or \dagger\dagger
   \or \ddagger\ddagger \else\@ctrerr\fi}}
\makeatother

\usepackage{eccvabbrv}

\usepackage{graphicx}
\usepackage{booktabs}
\usepackage{multirow}
\usepackage{tcolorbox}
\usepackage{float}
\usepackage{subcaption}
\usepackage{xspace}
\usepackage{wrapfig}

\newcommand{\myparagraph}[1]{\noindent \textbf{#1}}
\newcommand{\method}{TexTailor\xspace}

\usepackage[accsupp]{axessibility}  

\usepackage{hyperref}

\usepackage{orcidlink}

\begin{document}

\title{TexTailor: Inference-Time Textual Guidance Tailoring for Multimodal Diffusion Transformers}

\titlerunning{TexTailor}

\author{Binglei Li\inst{1,2}\orcidlink{0009-0004-1182-9348} \and
Mengping Yang\inst{1,3}\thanks{Project Lead.}\orcidlink{0000-0003-1503-9621} \and
Zhiyu Tan\inst{1,3}\orcidlink{0009-0000-8110-6066} \and \\
Junping Zhang\inst{1}$^*$\orcidlink{0000-0002-5924-3360} \and
Hao Li\inst{1,2,3}\thanks{H.~Li and J.~Zhang are the corresponding authors.}\orcidlink{0000-0001-6197-0674}
}

\authorrunning{B.~Li et al.}

\institute{Fudan University, Shanghai 200433, China \and
Shanghai Innovation Institute, Shanghai 200030, China \and
Shanghai Academy of AI for Science, Shanghai 200030, China\\
\email{\{blli24, zytan24\}@m.fdu.edu.cn,\\ \{mengpingyang, jpzhang, lihao\_lh\}@fdu.edu.cn}
}

\maketitle

\begin{figure}[htbp]
    \vspace{-15pt}
    \centering
    \includegraphics[width=\textwidth]{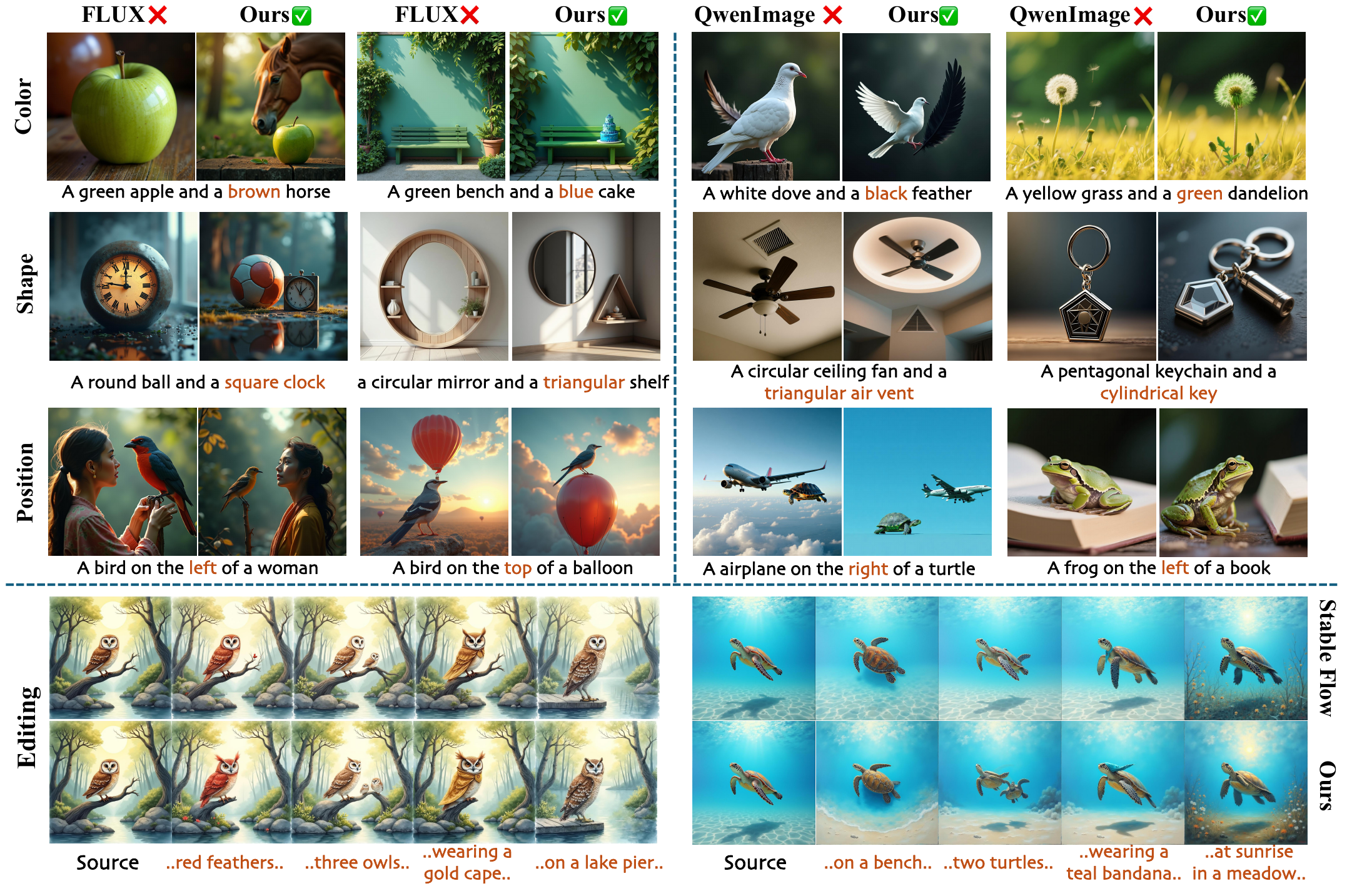} %
    \caption{
    \textbf{Visual comparison of text-to-image and editing results}, demonstrating that our \method yields better text alignment across semantic attributes.}
    \label{fig:teaser}
    \vspace{-20pt}
\end{figure}

\begin{abstract}

Recent breakthroughs of transformer-based diffusion models, particularly with Multimodal Diffusion Transformers (MMDiT) driven models like FLUX and Qwen Image, have facilitated thrilling experiences in visual generation.
However, these models rely only on the interactions between textual conditions and visual features to produce semantically aligned images. 
Once the interactions fail to reflect the nuanced compositional structure of the prompt, the generated images might be unsatisfactory.
Thus, a comprehensive understanding of how different blocks and their interactions with textual conditions is crucial for better understanding the intrinsic attributes and for enhancing their interactions accordingly to strengthen the prompts adherence.
In this paper, we first develop a systematic pipeline to comprehensively investigate each block's functionality by \textit{removing}, \textit{disabling}, and \textit{enhancing} textual hidden-states at corresponding blocks.
Our analysis reveals that 1) semantic information appears in earlier blocks and finer details are rendered in later blocks, 2) removing specific blocks is usually less disruptive than disabling text conditions, and 3) enhancing textual conditions in selective blocks improves semantic attributes.
Building on these observations, we propose \method, a novel inference-time method for tailoring block-wise textual guidance. Our approach not only improves text-image alignment but also enables a range of downstream applications, including precise editing and inference acceleration.
%
%
Extensive experiments demonstrated that our method outperforms various baselines and remains flexible across text-to-image generation, image editing, and inference acceleration.
%
Our method improves T2I-Combench from 56.92\% to 63.00\% and GenEval from 66.42\% to 71.63\% on SD3.5, without sacrificing synthesis quality. These results advance understanding of MMDiT models and provide valuable insights to unlock new possibilities for further improvements.
Our code is available at \href{https://github.com/phil329/TexTailor}{https://github.com/phil329/TexTailor}.
\keywords{MMDiT \and Block-wise Analysis \and Inference-time Tailoring}

%
%

\end{abstract}

\section{Introduction}
\label{sec:intro}

Diffusion models~\cite{ddpm_ho2020denoising, ddim_song2020,sohl2015deep}, especially diffusion transformers (DiT)~\cite{dit_AdaLN_peebles2023scalable, bao2023all}, have become the de-facto paradigm for real-world applications across various domains, including text-to-image~\cite{rombach2022high, chen2023pixart}, controllable generation~\cite{controlnet_zhang2023,tan2024ominicontrol} with depth~\cite{he2024litegfm,yang2024depth} and canny input, as well as diverse video generation applications~\cite{yang2024cogvideox, wan2025wan,yang2026omni}, unlocking unprecedented experiences for AI generated content.
In particular, the most recent approaches such as Stable Diffusion 3~\cite{SD3_esser2024scaling}, FLUX~\cite{flux_2024}, and Qwen-Image~\cite{qwenimage_wu2025technicalreport} further advance the synthesis quality via incorporating the flow matching~\cite{flowmatching_lipman2022, RectifiedFlow_liuflow} training objective and the top-performing multimodal diffusion transformer (MMDiT) architecture~\cite{SD3_esser2024scaling, flux_2024}.
Specifically, MMDiT concatenates vision and textual tokens and performs joint self-attention to facilitate a seamless textual-visual interaction between these modalities.
%

\begin{figure}[t]
  \centering
  \includegraphics[width=\linewidth]{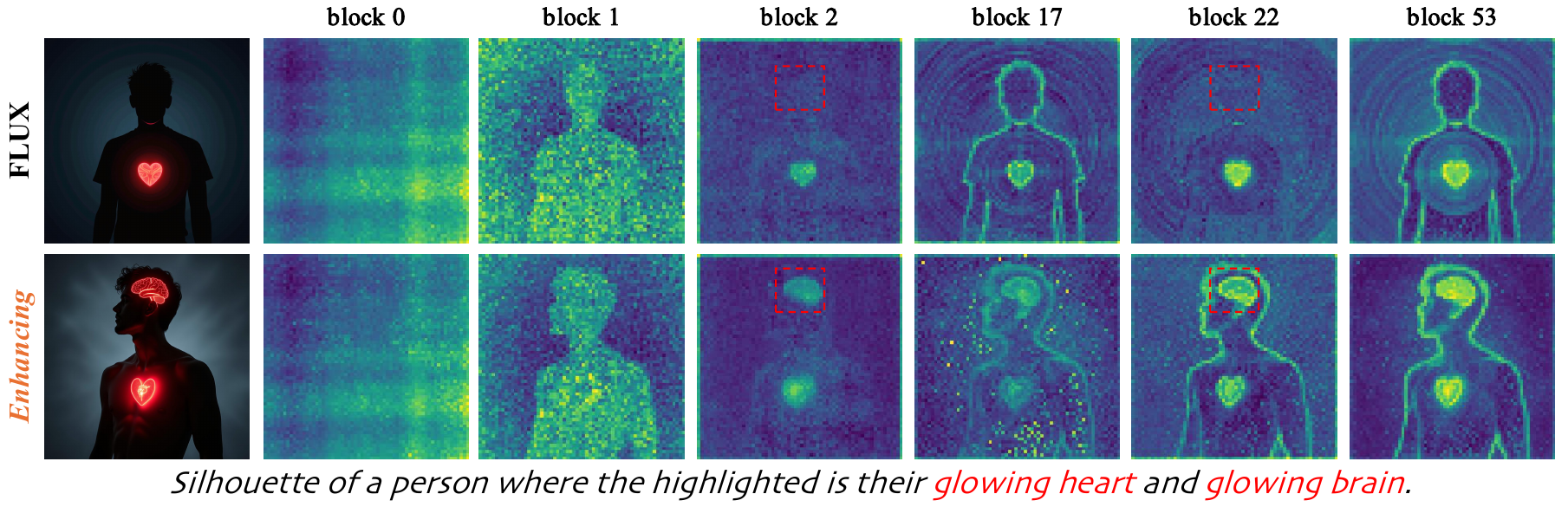}
  \caption{
  \textbf{Textual-Visual interaction attention maps across different blocks}.
  %
  }
  \label{fig:attnmap_showcase}
\end{figure}

%
Despite their remarkable success, MMDiT-based models produce images based on interactions between textual representations and visual hidden states only.
As shown in Fig.~\ref{fig:attnmap_showcase}, if the textual-visual interactions fail to faithfully render semantic details of given instructions, the generated images might be unsatisfactory.
Therefore, it is crucial to investigate the intrinsic mechanisms within MMDiT-based models, enabling better understanding on the internal interactions of MMDiT.
Following this philosophy, Stable Flow~\cite{stableflow_avrahami2025} detected vital blocks by bypassing each block, and TACA~\cite{taca_lv2025rethinking} proposed a timestep-aware attention weighting mechanism to balance multimodal interactions.
FreeFlux~\cite{freeflux_wei2025} and E-MMDiT~\cite{shin2025exploring} analyzed MMDiT's attention mechanism by shifting RoPE and decomposing attention metrics, respectively.
Semantic Routing~\cite{li2026semantic} and HyperAlign~\cite{xie2026hyperalign} respectively employ linear fusion and hypernetwork-generated low-rank adaptation weights to modulate text-visual interactions within diffusion transformers.
However, existing studies primarily focus on isolating or manipulating individual aspects, overlooking the synergistic effects that arise from the complex interactions across different blocks and modalities.
That is, it remains unclear how different internal MMDiT blocks interact with textual representations and how they collaborate with each other to produce high-quality and coherent outputs.
Consequently, a deeper and more detailed analysis of understanding how MMDiT blocks collectively contribute to rendering sophisticated outputs would not only enrich our understanding of MMDiT models but also open avenues for refining their synthesis quality and inference efficiency.
For instance, by identifying which blocks control specific attributes (\emph{e.g.,} color, shape, spatial relationships), we can revise the corresponding blocks accordingly (as illustrated from the results in Fig.~\ref{fig:teaser}).

%

To identify each block's detailed role and functionality, we first develop a systematic automatic pipeline to probe the internal cooperation of MMDiT blocks and their interactions with text conditions.
Our pipeline involves two main stages:
For the first stage, we construct a dedicated prompt set covering representative attributes including color, shape, and spatial relationship. These prompts are used to generate images from three popular MMDiT-based models (SD3.5, FLUX, Qwen Image).
Then,  we quantify the influence of block-wise interactions on the output images with perceptual (\emph{i.e.,} DINOv2~\cite{dinov2_oquab}) and semantic scores (\emph{i.e.,} CLIP Score~\cite{CLIP_radford2021learning}), as well as VQA results from general MLLMs (\emph{i.e.,} Qwen-VL~\cite{qwen25VL_bai2025})
Specifically, we probe block-wise interactions and collaboration via: 
1) \textit{removing} specific blocks to assess their individual contributions; 2) \textit{disabling} block-level textual conditions to test semantic understanding; and 3) \textit{enhancing} textual hidden-states of different blocks to investigate their potential to refine the coherence and detail of synthesized outputs.
Through these analysis, we reveal \emph{several insightful findings}:
First, semantic information appears in earlier blocks, while fine-grained details are rendered in later blocks.
As illustrated by the enlarged details in Fig.~\ref{fig:analysis_overall}, different attributes exhibit similar behaviors across various blocks, suggesting that a unified enhancement strategy can be applied to all attributes.
Second, removing certain blocks is less disruptive than disabling corresponding textual conditions, indicating MMDiT-based models' stronger reliance on conditional guidance and their robustness to removing individual blocks.
Last but not least, enhancing textual representations of selective blocks could improve overall text alignment without compromising synthesis quality.
These insights provide a better understanding towards the efficacy and interactions of different components within the MMDiT architecture, guiding further optimization and improvements in various applications.
Capitalizing on these observations, we propose \method, a novel inference time textual guidance enhancement framework to improve the text alignment of text-to-image, facilitate image editing, and accelerate model inference within MMDiT-based models.
Concretely, after identifying each block's contribution to the specific semantic attributes of output images, we can strategically enhance the text-visual interactions of these blocks to optimize the text alignment.
Moreover, a token-level tailoring strategy is further designed to improve the interactions of key textual tokens of specific attributes.
As shown in Fig.~\ref{fig:attnmap_showcase}, the textual-visual interactions after enhancing present improved adherence of given prompts, leading to more satisfactory output.
%
Regarding editing tasks, we can similarly attach higher importance to specific blocks that manage certain semantic attributes, such as ``color'' and ``count'', ensuring that edits are accurately and effectively performed.
%
Additionally, we could accelerate the inference process by skipping blocks that are less critical for semantic understanding, thus streamlining computations while preserving synthesis quality.
Together, our framework facilitates efficient, precise, and generalizable model performance across different tasks without requiring additional training.
Extensive results demonstrate that our method significantly advances the model performance across various baselines (SD3.5, FLUX, and Qwen-Image), benchmarks (GenEval~\cite{geneval_ghosh2023}, T2I-Combench~\cite{t2i_compbench_huang2023t2i}) and metrics (CLIP-Score~\cite{CLIP_radford2021learning}, Human-Preference Score~\cite{HPSV2_wu2023human}, Aesthetic Score~\cite{laion_aesthetic_2022}), as well as different tasks (generation, editing, and acceleration), consistently showing the effectiveness and generalizability of our method.
%
%

We summarize our primary contributions as:
1) \textbf{A systematic framework to probe block-wise interactions of MMDiT-based models}. We systematically investigate the internal interactions across blocks and modalities within several MMDiT-based models, offering valuable and consistent insights to guide further improvements.
2) \textbf{A novel inference-time framework for tailoring textual guidance}. We propose \method to enhance the interactions between textual conditions and visual features within different blocks, thereby improving the text alignment of generated images.
Moreover, our method can be applied to other applications including precise editing and inference acceleration, fully unlocking the potential of baseline models.
3) \textbf{Extensive evaluations across multiple baseline models and diverse benchmarks} for various tasks consistently demonstrate the effectiveness and generalizability of our approach in advancing model performance.

\section{Related Work}

\label{sec:related}

\myparagraph{Diffusion Transformers.}
Diffusion Transformers~\cite{dit_AdaLN_peebles2023scalable} have become the dominant paradigm for high-fidelity image and video generation, which adopt transformer~\cite{attention_vaswani2017} architecture as the main backbone, demonstrating superior scalability and training efficiency compared to previous UNet-based~\cite{ddpm_ho2020denoising,diffusion_beat_gan_dhariwal2021,cfg_ho2022classifier} models.
Recent variants, such as open-sourced SD3~\cite{SD3_esser2024scaling}, FLUX~\cite{flux_2024}, Qwen Image~\cite{qwenimage_wu2025technicalreport}, Hunyuan Image~\cite{cao2025hunyuanimage} and Hunyuan Video~\cite{wu2025hunyuanvideo}, and commercial models like Seedream series~\cite{seedream2_gong2025,seedream3_gao2025}, Sora~\cite{OpenAI2024_Sora}, Imagen3~\cite{imagen3_baldridge2024}, further advance text-to-image/video to an unprecedented level with the top-performing multimodal diffusion transformer (MMDiT) architecture.
Besides scaling the MMDiT-based models, many efforts have also focused on accelerating the iterative denoising process~\cite{dpmsolver_lu2022,CM_song2023consistency,LCM_luo2023latent}, controlling the results~\cite{controlnet_zhang2023,tan2024ominicontrol,instanceassemble_xiang2025}, editing the outputs~\cite{stableflow_avrahami2025,freeflux_wei2025}, \emph{etc.}

\myparagraph{Understanding and Improving Diffusion Models.}
Numerous prior works proposed various techniques to analyze the roles of different components of UNet-based diffusion models.
For instance, P2P~\cite{hertz2022prompt} showed that cross-attention layers are essential for rendering the spatial layout,  MasaCtrl~\cite{cao2023masactrl} and~\cite{liu2024towards} demonstrated that self-attention are more important for preserving the geometric and shape details.
FreeU~\cite{si2024freeu} and PBC~\cite{zhou2025exploring} respectively analyzed the functionality of skip connections and position encoding mechanism in diffusion UNet.
Further, Yi \etal~\cite{yi2024towards} investigated the mechanism of text prompts and Williams \etal~\cite{williams2023unified} developed a unified framework for analysing the UNet architectures.
P+~\cite{voynov2023p} optimizes extended per-layer text embeddings to improve prompt fidelity.
Attend-and-Excite~\cite{chefer2023attend} amplifies cross-attention maps for semantically relevant tokens at inference time.
However, the understanding of different components within MMDiT-based models remains underexplored and it is crucial to gain a holistic understanding of these components to advance the field. 
Existing arts tried to identify the contribution of different layers~\cite{stableflow_avrahami2025}, rotary position embeddings (RoPE)~\cite{freeflux_wei2025}, and attention~\cite{shin2025exploring}, yet they usually focus on specific applications like editing and lack a systematic analysis of the internal interplay of components within models.
TACA~\cite{taca_lv2025rethinking} indicated an imbalanced issue in the cross-model attention and ameliorated it with a timestep-aware re-weighting scheme.
Semantic Routing~\cite{li2026semantic} explored the multi-layer LLM feature weighting for diffusion transformers, which trains a linear fusion module to modulate the text-visual interactions.
HyperAlign~\cite{xie2026hyperalign} proposed a hypernetwork to generate LoRA weights to modulate the diffusion model's generation operators.
Nevertheless, none of the approaches provides a holistic view of how these components jointly influence the model's performance.

\section{Systematic Analysis of Block-wise Interactions}
\label{sec:analysis}

\subsection{Preliminaries}
\label{sec:pre}

\myparagraph{Diffusion Models.} 
Diffusion models (DMs) involve a forward process and a reverse generation process.
During the forward process, random noises are gradually added to data ($\mathbf{x}_0\sim q(x)$) across $t \sim (1...T)$  timesteps:
\begin{equation}
    \mathbf{x}_t = \sqrt{\alpha_t}\mathbf{x}_{t-1} + \sqrt{1 - \alpha_t}\epsilon_{t-1}.
\end{equation}
In the reverse generation process, the model iteratively reconstruct the original data following a trajectory opposite to  the forward process:
\begin{equation}
p_\theta(\mathbf{x}_{t-1}|\mathbf{x}_t) = \mathcal{N}(\mathbf{x}_{t-1}; \mu_\theta(\mathbf{x}_t, t), \Sigma_\theta(\mathbf{x}_t, t)),
\end{equation}
where $\mu_\theta$ and $\Sigma_\theta$ are learnable mean and covariance, respectively.

\myparagraph{MMDiT-based Models.}
MMDiT-based models, pioneered in SD3~\cite{SD3_esser2024scaling}, leverage a joint multimodal architecture to process text embeddings $\boldsymbol{c} \in \mathbb{R}^{N_{\boldsymbol{c}} \times D}$ and visual features $\boldsymbol{x} \in \mathbb{R}^{N_{\boldsymbol{x}} \times D}$ in a unified attention operation by concatenating them as $h_{in} = [\boldsymbol{c};\boldsymbol{x}] \in \mathbb{R}^{(N_{\boldsymbol{c}} + N_{\boldsymbol{x}}) \times D}$.
This sequence is then processed by multiple MMDiT blocks with a joint self-attention layer:
\begin{equation}
    \begin{aligned}
        \text{Attention}(Q, K, V) = {softmax}({QK^T}/{\sqrt{d_k}})V,
    \end{aligned}
\end{equation}
where $Q, K, V$ denote the concatenated query, key and value of text and image tokens.
The MMDiT blocks serve as the core modules enabling effective and fine-grained integration of textual-visual information during denoising.

\subsection{Understanding Block-wise Interactions of MMDiT}
\label{sec:block_importance}

\begin{figure}[t]
    \centering
    \includegraphics[width=\linewidth]{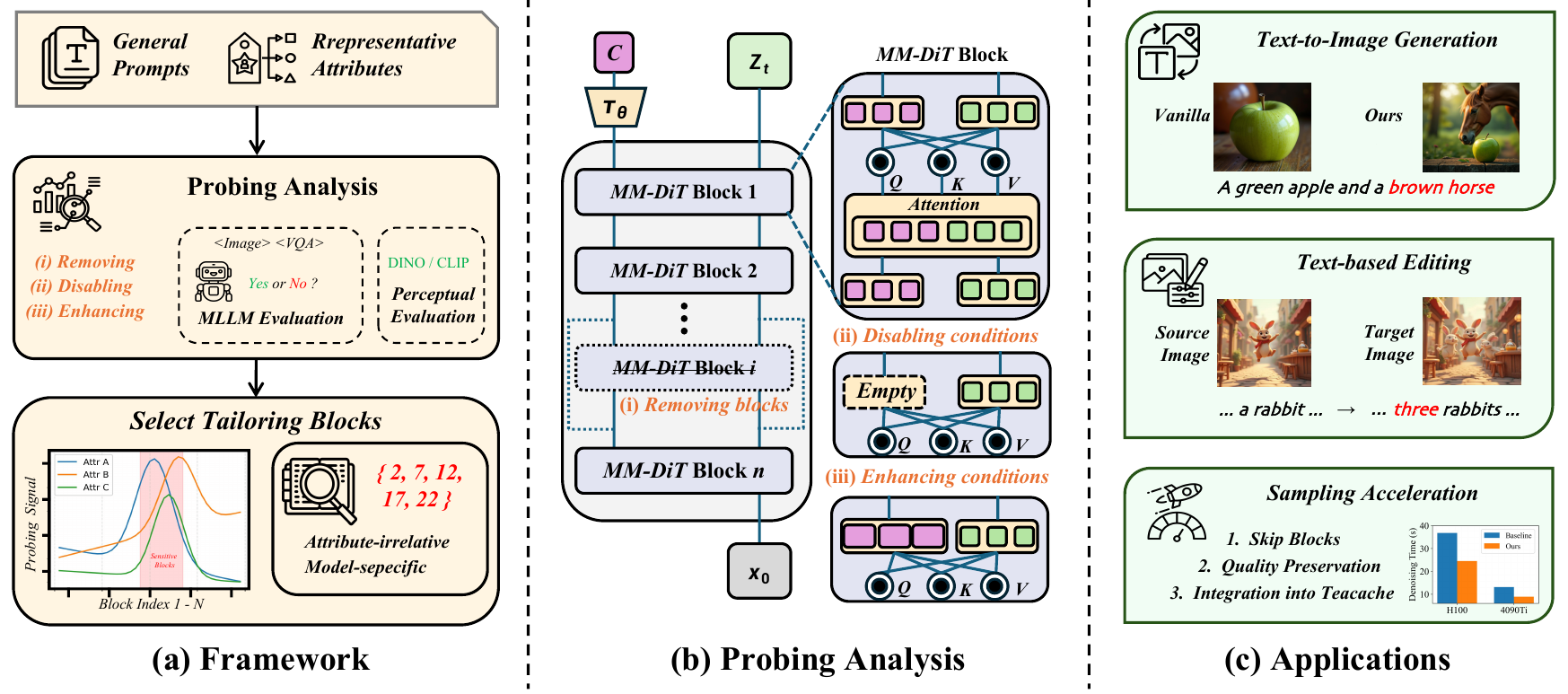}
    \caption{
    \textbf{Analysis and framework overview.}
    (a) We propose \method, a training-free framework to tailor the inference-time text-visual interactions.
    (b) Our probing analysis consists of three key operations: \textit{removing}, \textit{disabling}, and \textit{enhancing}.
    %
    The tailoring blocks are chosen from the high-signal region, which can be applied to various downstream tasks, including T2I generation, image editing and acceleration (c).
    }
    \label{fig:analysis_overview}
\end{figure}

%
In this part, we develop a systematic framework to automatically investigate block-wise interactions and their influence on the representative semantic attributes (\emph{i.e.,} color, shape, spatial relationships).
As shown in Fig.~\ref{fig:analysis_overview} (b), our study involves three key operations:
1) \textit{removing} specific blocks to probe each block's individual contribution;
2) \textit{disabling} textual conditions of different blocks to evaluate their reliance on semantic guidance; 
3) \textit{enhancing} textual guidance of certain blocks to investigate their potential to refine coherence and detail.
Specifically,  we amplify the text condition hidden states by a factor of $2$ as $\boldsymbol{c} \rightarrow 2\boldsymbol{c}$, to investigate each block's latent capacity for assimilating semantic information.
Regarding the disabling operation, we mute the textual hidden states via attaching an empty tensor with $torch.zeros\_like(c)$.
We expect that the probing analysis could generalize to various attributes and categories, so we construct a prompt dataset from the widely adopted HPDv2~\cite{HPSV2_wu2023human} and COCO~\cite{lin2014microsoftcoco} datasets.
It comprises $633$ diverse prompts across three representative attributes: ``color'', ``shape'', and ``spatial relationships''.
For each prompt, we perform \textit{removing}, \textit{disabling} and \textit{enhancing} on SD3.5-Large~\cite{SD35_2024}, FLUX.1-Dev~\cite{flux_2024}, and Qwen Image~\cite{qwenimage_wu2025technicalreport} models, operating on one block at a time.
For the probing analysis, we evaluate generated images using Qwen2.5-VL-72B~\cite{qwen25VL_bai2025} via question-answering pairs on the prompts and the generated images.
Further, we evaluate perceptual (DINOv2~\cite{dinov2_oquab}) and semantic similarities (CLIP Score~\cite{CLIP_radford2021learning})  between images from our modified and original models to quantify the effect of our block-wise manipulations.
Notably, despite the limited number of prompts per attribute, repeated sampling with $5$ fixed seeds produces consistent and reliable results (see supplementary Sec.~\ref{appendix:probing_implementation_details} for more details).

\begin{figure}[t]
    \centering
    \begin{subfigure}{0.49\textwidth}
        \centering
        \includegraphics[page=1,width=\linewidth]{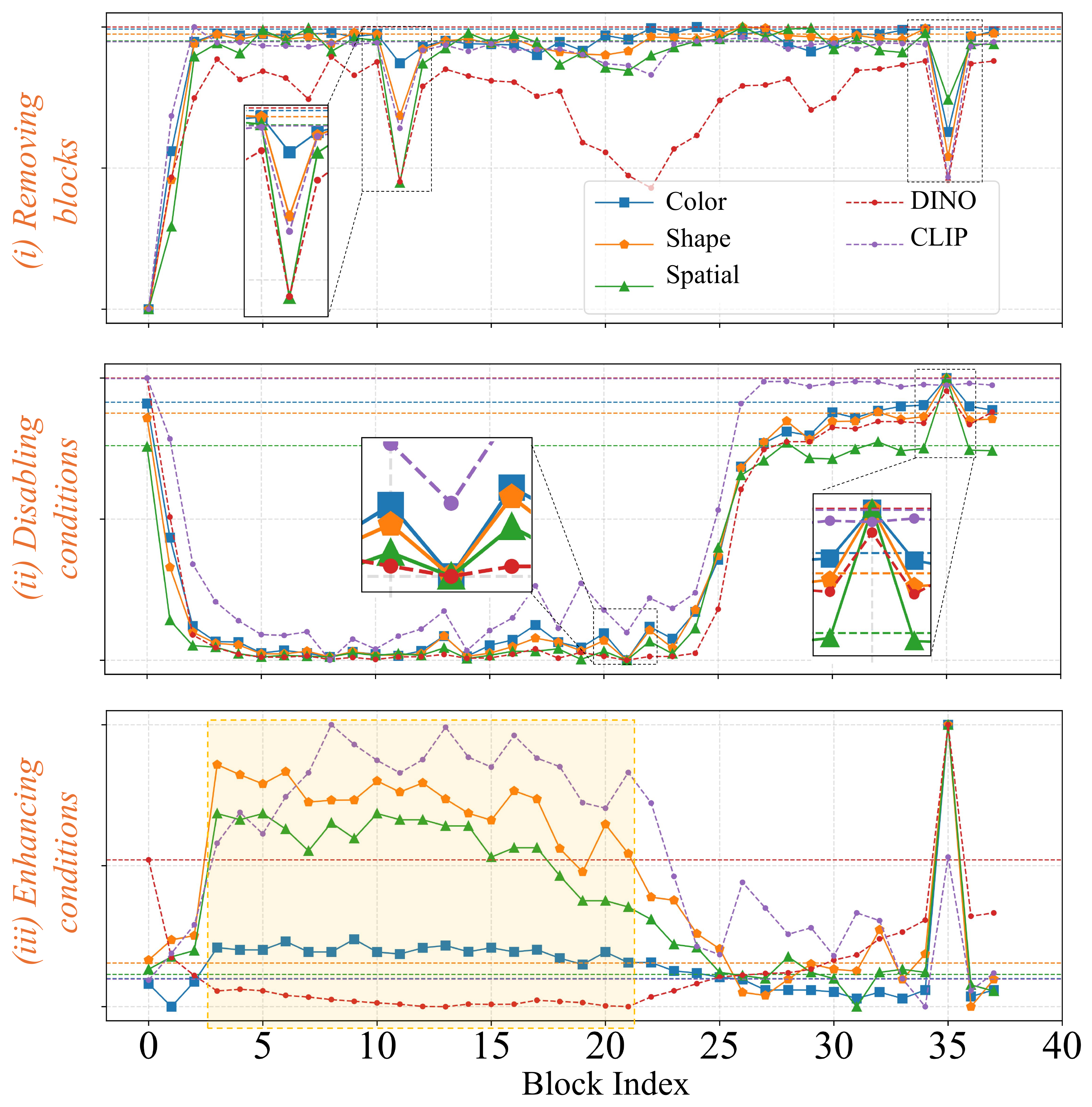}
        \caption{Probing analysis of SD3.5-Large.}
        \label{fig:analysis_sd35}
    \end{subfigure}
    \hfill
    \begin{subfigure}{0.49\textwidth}
        \centering
        \includegraphics[page=2,width=\linewidth]{figures/probing_plot.pdf}
        \caption{Probing analysis of FLUX.1-Dev.}
        \label{fig:analysis_flux}
    \end{subfigure}

    \caption{
    \textbf{Block-wise analysis results across various MMDiT-based models on representative attributes}.
    We observe that different attributes exhibit similar block-wise behaviors, which indicates that the probing analysis is effective and generalizable.
    %
    %
    %
    We select tailoring blocks from the common high-signal region (yellow box) for all attributes.
    Zoom in for details and see supplementary for the results of Qwen Image.
    }
    \label{fig:analysis_overall}
\end{figure}

The analysis results on the representative attributes across SD3.5 ($38$ blocks), FLUX ($57$ blocks) are shown in Fig.~\ref{fig:analysis_overall}.
For each subfigure, we plot the DINOv2 and CLIP score curves, along with the quantitative curves of performing our analysis method, \emph{i.e.,} removing (1st row), disabling (2nd row), enhancing (3rd row), on three representative attributes.
Despite testing on different models, we could consistently observe several interesting observations from these results.
%

\myparagraph{Observations 1: Removing less critical blocks tends not to significantly impact overall performance.}
Fig.~\ref{fig:analysis_sd35} (top) and Fig.~\ref{fig:analysis_flux} (top) illustrate the impact of \textit{removing} different blocks.
We could observe that all models are sensitive to the removal of the earlier ($0-5$) and late blocks, leading to a significant drop in synthesis performance.
We attribute this sensitivity to the critical roles of early blocks in initializing inputs and of late blocks in refining details for the final output.
By contrast, removing blocks from the middle layers generally results in a smaller impact, as also evidenced by both the synthesis performance and the DinoV2 and CLIP scores.
Such observation indicates that these blocks might be less critical for maintaining the fidelity and coherence of the generated outputs. 
Accordingly, we could remove some blocks to optimize model efficiency without compromising the synthesis quality.
Furthermore, we find that removing some blocks (the $18th$ block in FLUX, the $35th$ block in SD3.5) leads to a significant drop in performance, which may be the boundary between the dual-stream and single-stream blocks or the vital blocks for the semantic rendering.
%

\myparagraph{Observations 2: Disabling textual conditions is more disruptive than removing specific blocks.}
Fig.~\ref{fig:analysis_sd35} (middle) and Fig.~\ref{fig:analysis_flux} (middle) reveal that disabling textual conditions, especially in the earlier blocks ($0-20$), causes a more pronounced degradation in the synthesis performance compared to merely removing specific blocks.
That is, textual conditions play a crucial role in guiding the models' generative process.
Moreover, we can see that the results of disabling conditions of late blocks are less detrimental to the overall performance, particularly the CLIP Score, suggesting that these blocks are specialized in refining details and the core semantics are rendered by the earlier blocks.
Interestingly, we observe that \textit{disabling} in the dual-stream part of FLUX causes larger degradation. In contrast, Qwen Image is more robust, whose performance is mainly affected by the first and last few layers, while disruptions in intermediate layers can be mitigated by later modules.

\myparagraph{Observations 3: Enhancing textual conditions on certain blocks could improve the synthesis performance.}
Fig.~\ref{fig:analysis_sd35} (bottom) and Fig.~\ref{fig:analysis_flux} (bottom) show the enhancing results.
Though the simple $\times2$ operation may not yield optimal results, enhancing textual conditions on certain blocks can improve the synthesis performance.
Remarkably, for all three attributes, all models show performance improvements compared with the original baseline, despite Qwen Image showing less improvement due to its strong baseline.
Interestingly, CLIP Score shows the same trend as the attributes, indicating that enhancing textual conditions on certain blocks could improve the text-visual alignment.
To our knowledge, this observation has never been documented in existing literature.
We highlight the blocks in the yellow region that all the representative attributes commonly exhibit significant performance improvements, which are the tailoring block ranges for each attribute in the latter sections.
In return, one could manipulate specific attributes (e.g., ``color'', ``shape'', ``count'' and ``position'' in Fig.~\ref{fig:teaser} and Fig.~\ref{fig:editing_qualitative}) by altering textual information at the selected tailoring blocks.
%
%


Overall, our analysis provides a comprehensive investigation of the block-wise capabilities and their interactions with textual conditions, yielding several interesting insights on how different blocks contribute to the output.
These findings contribute to a better understanding of MMDiT-based models, offering valuable perspectives that could facilitate further enhancements and optimizations.

\section{Methodology}

\label{sec:method}

\subsection{Motivations}
%
%
The diffusion models primarily rely on the text prompts to guide the generation process.
Attention mechanisms in MMDiT models~\cite{SD3_esser2024scaling,flux_2024,qwenimage_wu2025technicalreport} are designed to capture the interactions between text and image features, which are crucial for the generation process.
Fig.~\ref{fig:attnmap_showcase} presents attention maps from FLUX and our enhanced results, where our method strengthens attention to the relevant text phrase ``glowing brain'', improving text-visual alignment.
%
%
%
Moreover, different blocks play different roles in the generation process, and our enhancing method can help the model to focus on the relevant text tokens.
Our analysis demonstrates that tailoring only a few specific blocks is sufficient to strengthen the text-image interactions. 
This tailoring strategy not only effectively modulates the attention mechanism, but also mitigates the problem of imbalanced visual and textual tokens~\cite{taca_lv2025rethinking}.
%
%
Our work offers the community a fresh perspective and actionable findings that go beyond the traditional block-wise analysis.
Although conducted on the three representative attributes, the tailoring blocks could generalize to other attributes.
%

\subsection{\method: Tailoring Textual Guidance for Denoising}
%
We propose a straightforward, training-free method to tailor text-visual interactions within blocks by capitalizing on their pivotal roles, which is highlighted in the yellow boxes in Fig.~\ref{fig:analysis_overall}(bottom panel).
Specifically, we enhance the hidden states of textual conditions in these vital blocks $\mathcal{S}=\{s_1, s_2, \ldots, s_m\}$ by a factor of $\lambda(l)$ using the following equation:
\begin{equation}
  \label{eq:enhance_token}
    \boldsymbol{c}_{enh}^{(l)} =(1 - M) \odot \boldsymbol{c}^{(l)} +\lambda(l)\cdot M \odot  \boldsymbol{c}^{(l)},\quad \forall l \in \mathcal{S}.
  \end{equation}
where $\boldsymbol{c}^{(l)}$ is the original textual hidden states of block $l$ and $\odot$ is element-wise multiplication. $\lambda(l)$ can be a constant or a block-dependent function.
$M$ is a binary mask that denotes corresponding indices of enhanced textual tokens.

\myparagraph{Full sentence tailoring.}
Formally, we set $M$ to be a full one vector to tailor the entire textual condition.
The equation becomes $\boldsymbol{c}_{enh}^{(l)} =\lambda(l) \odot  \boldsymbol{c}^{(l)}$, which is suitable for all prompts.
Full sentence tailoring can avoid the problem of difficulty locating the specific token indices, such as complex prompts containing multiple attributes, which cannot obtain an accurate mask $M$ for token-level tailoring.

\begin{figure}[t]
  \centering
  \includegraphics[width=\linewidth]{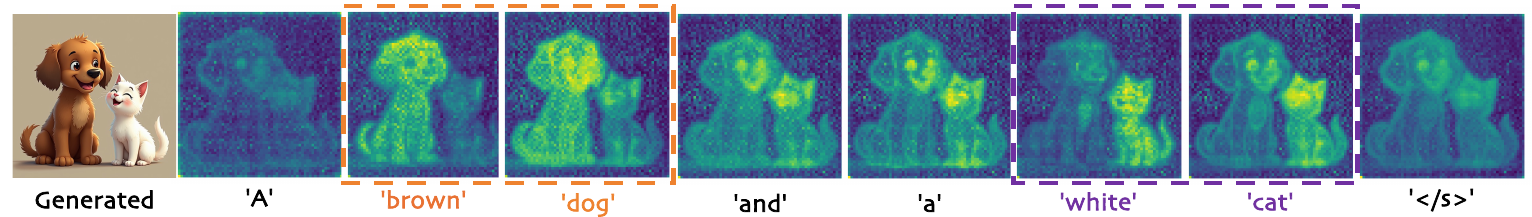}
  \caption{
  \textbf{Attention map of different tokens}.
  The target phrases well match the attention map of the corresponding tokens, highlighted in dashed boxes.
  We can enhance specific tokens to boost their impact on the attention map.
  }
  \label{fig:token_atten}
\end{figure}

\myparagraph{Token-level tailoring.}
To further improve the semantic understanding capability of certain blocks on specific attributes, we introduce token-level enhancement to amplify key textual tokens.
As shown in Fig.~\ref{fig:token_atten}, we can determine the mask $M$ by locating the token indices of the target phrases within the tokenized prompt.
Such an operation ensures that critical semantic attributes receive greater emphasis.
Specifically, some attributes, such as ``color'', ``shape'', and ``count'', are easy to locate the corresponding mask $M$, on which we conduct token-level tailoring experiments.
Thus, this allows for a better understanding of textual conditions, emphasizing key semantic attributes within the model.

\section{Experiments}
\label{sec:exp}
\label{sec:exp_implementation_details}

\subsection{Experiment Setups}
\label{sec:exp_implementation_details}
%
\myparagraph{Implementation Details.}
We apply our method on several state-of-the-art MMDiT-based models, namely SD3.5-Large~\cite{SD35_2024}, FLUX.1-Dev~\cite{flux_2024} and Qwen Image~\cite{qwenimage_wu2025technicalreport}.
Based on the observations from our systematic analysis in Sec.~\ref{sec:analysis}, we tailor the textual conditions during denoising on the selected blocks according to the attribute in Tab.~\ref{tab:enhance_blocks}.
%
%
%
%
%
The tailoring parameter $\lambda(l)$ in Eq.~\ref{eq:enhance_token} is set to $1.5$ unless otherwise specified.
We compare our method on text-to-image generation with TACA~\cite{taca_lv2025rethinking}, which attaches a timestep-aware importance on the textual conditions within the attention.
We report the full sentence tailoring and token-level tailoring results on the T2I-CompBench and GenEval benchmarks.
During inference, we sample $1024\times 1024$ images using the EDM~\cite{karras2022elucidating} sampler for $28$ denoising steps, and use the official default settings for the other inference parameters.

\myparagraph{Evaluation Metrics.}
We evaluate our method on the T2I-CompBench~\cite{t2i_compbench_plusplus_huang2025t2i} and Geneval~\cite{geneval_ghosh2023} benchmarks, which are widely adopted for text-to-image alignment.
Additionally, we utilize Aesthetics score~\cite{laion_aesthetic_2022} and HPSv2~\cite{HPSV2_wu2023human} to evaluate the overall image quality, ensuring that our method does not negatively impact the synthesis quality of the original models.
We follow the official guidance for evaluation.
All experiments are carried out on NVIDIA 4090 and H100 GPUs.

\myparagraph{Selected Blocks for Tailoring.}
\label{sec:exp_selected_block_for_tailoring}
%
In the Sec.~\ref{sec:analysis}, we systematically investigate the block-wise interactions in MMDiT-based models, and identify similar patterns across three representative attributes.
In the Fig.~\ref{fig:analysis_overall}, the representative attributes exhibit similar block-wise behaviors, as highlighted in the enlarged details.
This indicates that the selected tailoring blocks can be generalized to other attributes.
We uniformly select the proper number of blocks, $4$ for SD3.5-Large and $5$ for FLUX.1-Dev and Qwen Image, respectively, for tailoring in the yellow region of Fig.~\ref{fig:analysis_overall}(bottom panel).
Tab.~\ref{tab:enhance_blocks} shows the selected tailoring blocks for different models, on which we conduct the tailoring experiments.

\begin{table}[t]
    \centering
    \scriptsize
    \caption{
    \textbf{Blocks for tailoring the text-visual interactions of different models.}
    }
    \label{tab:enhance_blocks}
    \setlength{\tabcolsep}{14pt}
    \begin{tabular}{lccc}
    \toprule
    Models & SD3.5-Large& FLUX.1-Dev & Qwen Image \\
    \midrule
    Total Blocks & 38 & 57 & 60 \\
    Tailoring Blocks & \textbf{\{3,9,15,21\}} & \textbf{\{2,7,12,17,22\}} & \textbf{\{3,9,15,21,27\}} \\
    \bottomrule
    \end{tabular}
\end{table}

\subsection{Improved Text Alignment of Text-to-Image Generation}
\label{sec:exp_main_results}

\begin{figure}[t]
    \centering
    \includegraphics[width=\linewidth]{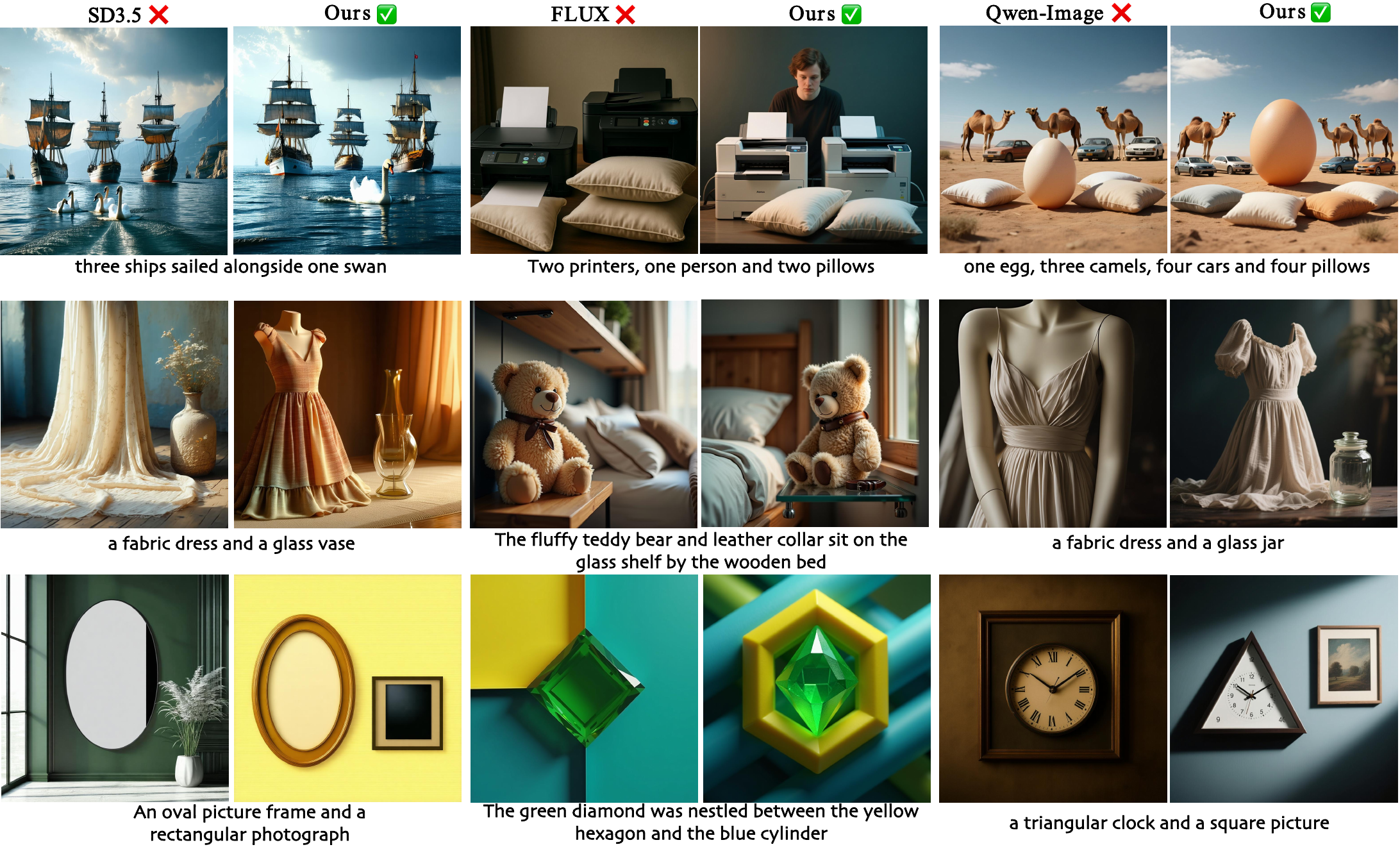} %
    \caption{
    \textbf{Qualitative comparisons between baselines and our method}. 
    Our method significantly improves the text alignment across various semantic attributes, including amount, colors, textures, and complex prompts, \emph{etc.}
    %
    %
    Zoom in for details. 
    }
    \label{fig:t2i_qualitative}

\end{figure}

\myparagraph{Qualitative Results.}
Fig.~\ref{fig:teaser} and Fig.~\ref{fig:t2i_qualitative} show the qualitative results of our \method compared to three baseline models.
\method significantly improves the text alignment across various semantic attributes, including amount, colors, textures, and complex prompts, \emph{etc}.
Even though the tailoring blocks are selected based on the representative attributes, the other attributes can also benefit from our method.
This observation demonstrates the generalizability of the tailoring blocks across different attributes.
See more results in Sec.~\ref{appendix:more_quantitative_results}.

\begin{table}[t]
    \centering
    \scriptsize
    \caption{
    \textbf{Quantitative results on T2I-CompBench with token-level and full sentence tailoring results}.
    \method consistently improves the text alignment on three models across various attributes.
    %
    }
    \label{tab:t2i_compbench}
    \setlength{\tabcolsep}{1.5pt}
    \begin{tabular}{@{}lcccccccc|cc@{}}
    \toprule
    \multicolumn{1}{c}{\multirow{2}{*}{Model}} &
    \multicolumn{3}{c}{Attribute Binding} &
    \multicolumn{3}{c}{Object Relationship} &
    \multicolumn{1}{c}{\multirow{2}{*}{Count}} &
    \multicolumn{1}{c}{\multirow{2}{*}{Comp.}} &
    \multicolumn{2}{|c}{Image Quality} \\ 
    
    \cmidrule(lr){2-4} \cmidrule(l){5-7} \cmidrule(l){10-11} 
    \multicolumn{1}{c}{} &
    \multicolumn{1}{c}{Color} &
    \multicolumn{1}{c}{Shape} &
    \multicolumn{1}{c}{Texture} &
    \multicolumn{1}{c}{2D Spa.} &
    \multicolumn{1}{c}{3D Spa.} &
    \multicolumn{1}{c}{NonSpa.} &
    \multicolumn{1}{c}{} &
    \multicolumn{1}{c}{} &
    \multicolumn{1}{|c}{HPSv2} &
    \multicolumn{1}{|c}{Aes.} \\
    
    \midrule
    SD3.5 & 0.7284 & 0.5592 & 0.7471 & 0.2866 & 0.3816 & 0.3118 & 0.5969 & 0.3727 & 29.2869 & 6.0978 \\
    TACA & 0.7434 & 0.5784 & 0.7444 & 0.2947 & 0.3839 & 0.3114 & 0.6029 & 0.3820 & \textbf{29.3225} & \textbf{6.2297} \\
    Ours(token) &  \textbf{0.8061}  &  \textbf{0.6789}  &  -  &  -  &  -  &  -  &  \textbf{0.6088}  &  -  &  29.0384  &  6.0346  \\
    Ours(full) & 0.8052 & 0.6744 & \textbf{0.8428} & \textbf{0.3647} & \textbf{0.3923} & \textbf{0.3169} & 0.5987 & \textbf{0.4047} & 28.9501 & 5.9401 \\

    \midrule
    FLUX & 0.7322 & 0.4908 & 0.6490 & 0.2935 & 0.3739 & 0.3044 & 0.5877 & 0.3597 & 29.1586 & 6.3563 \\
    TACA(r=64) & 0.7535 & 0.5126 & 0.6522 & 0.3043 & 0.3814 & 0.3045 & 0.5855 & 0.3619 & 29.1525 & 6.3327 \\
    TACA(r=16) & 0.7296 & 0.4898 & 0.6549 & 0.2991 & 0.3790 & 0.3034 & 0.5780 & 0.3585 & 29.1375 & 6.3205 \\ 
    Ours(token) &  \textbf{0.7815}  &  \textbf{0.5500}  &  -  &  -  &  -  &  -  &  \textbf{0.6091}  &  -  &  \textbf{29.2329}  &  \textbf{6.4119}  \\
    Ours(full) & 0.7804 & 0.5482 & \textbf{0.6980} & \textbf{0.3280} & \textbf{0.3900} & \textbf{0.3054} & 0.5860 & \textbf{0.3691} & 29.2267 & 6.4110 \\

    \midrule
    Qwen Image & 0.8554 & 0.6358 & 0.7650 & 0.3973 & 0.4077 & 0.3110 & 0.7406 &0.3983 & 28.8831 & 6.1925 \\
    Ours(token) &  \textbf{0.8688}  &  \textbf{0.6369}  &  -  &  -  &  -  &  -  &  \textbf{0.7616}  &  -  &  29.0204  &  6.2307  \\
    Ours(full) & 0.8677 & 0.6348 & \textbf{0.7796} & \textbf{0.4560} & \textbf{0.4202} & \textbf{0.3123} & 0.7359 & \textbf{0.4104} & \textbf{29.0212} & \textbf{6.2378} \\

    \bottomrule
    \end{tabular}
\end{table}

\begin{table}[t]
    \centering
    \scriptsize
    \setlength{\tabcolsep}{3.5pt}
    \caption{
    \textbf{Quantitative comparison on GenEval.}
    * denotes token-level tailoring.
    }
    \label{tab:geneval_result}
    \begin{tabular}{@{}lccccccc|cc@{}}
    \toprule
    Model & Overall & Single & Two & Count$^*$ & Color & Pos. & ColorAttr & HPSv2 & Aes. \\
    \midrule
    SD3.5  & 0.6642  & 0.9438 & 0.8939 & 0.6344  & 0.8059 & 0.2325 & 0.4750 & \textbf{29.5759} & \textbf{5.8871} \\
    + Ours(full)  & \textbf{0.7163}  & \textbf{0.9781} & \textbf{0.9672} & \textbf{0.6375}  & \textbf{0.8650} & \textbf{0.3925} & \textbf{0.4825} & 29.3729 & 5.7902 \\
    \midrule
    FLUX & 0.6538 & \textbf{0.9904} & 0.8258 & 0.6375  & 0.7713 & 0.2575 & 0.4400 & 29.8115 & 6.3650 \\
    + Ours(full) & \textbf{0.6826} & 0.9688 & \textbf{0.8914} & \textbf{0.6438}  & \textbf{0.7739} & \textbf{0.3475} & \textbf{0.4700} & \textbf{29.8207} & \textbf{6.4043} \\
    \midrule
    Qwen Image & 0.8551  & \textbf{0.9906} & 0.9520 & 0.8562  & 0.8617 & 0.7375 & 0.7325 & 30.4510 & \textbf{6.2327} \\
    + Ours(full) & \textbf{0.8777}  & \textbf{0.9906} & \textbf{0.9722} & \textbf{0.8594} & \textbf{0.8989} & \textbf{0.7475} & \textbf{0.7975} & \textbf{30.6851} & 6.2113 \\
    \bottomrule
    \end{tabular}%
\end{table}

\myparagraph{Quantitative Results.}
Tab.~\ref{tab:t2i_compbench} and Tab.~\ref{tab:geneval_result} show the token-level and full tailoring results on T2I-CompBench and GenEval benchmarks.
We could observe that our proposed method consistently obtains performance gains across various attributes on all three models, demonstrating the superiority and flexibility of our method.
Remarkably, we achieve substantial improvement of 12\% on Shape, 10\% on Texture, and 8\% on Color, in a totally training-free manner.
Although due to the inherent limitations of the baseline models in the attribute of quantity, our token-level tailoring performs better than default sentence tailoring, outperforming the baseline performance.
Additionally, the quantitative results of HPSv2 and Aesthetics scores demonstrate that our method improves the text alignment while maintaining the high aesthetic quality.
Together with the qualitative results in Fig.~\ref{fig:teaser} and Fig.~\ref{fig:t2i_qualitative}, the quantitative results further show that the proposed \method consistently improves semantic understanding in image generation tasks and generalizes well across various attributes.

\subsection{Ablation Analysis}
\label{sec:exp_ablation}

\begin{table}[t]
    \centering
    \scriptsize
    \caption{
    \textbf{Ablation analysis on smaller $\lambda$ and block selections}.
    }
    \setlength{\tabcolsep}{12pt}
    \begin{tabular}{lccc}
        \toprule
        Configs & Color & Shape & 2D Spatial \\
        \midrule
        $\lambda$ = 0.7  & 0.3161 & 0.2653 & 0.1100 \\
        $\lambda$ = 0.9  & 0.6891 & 0.4395 & 0.2611 \\
        \midrule
        Enhancing Random 5 blocks & 0.7624 & 0.5072 & 0.3119 \\
        Enhancing $All$ blocks      & 0.2360 & 0.2736 & 0.0495 \\
        Ours            & \textbf{0.7804} & \textbf{0.5482} & \textbf{0.3280} \\
        \bottomrule
    \end{tabular}
    \label{tab:ablate_blocks_strength}
\end{table}

\begin{figure}[t]
    \centering
    \includegraphics[width=0.9\textwidth]{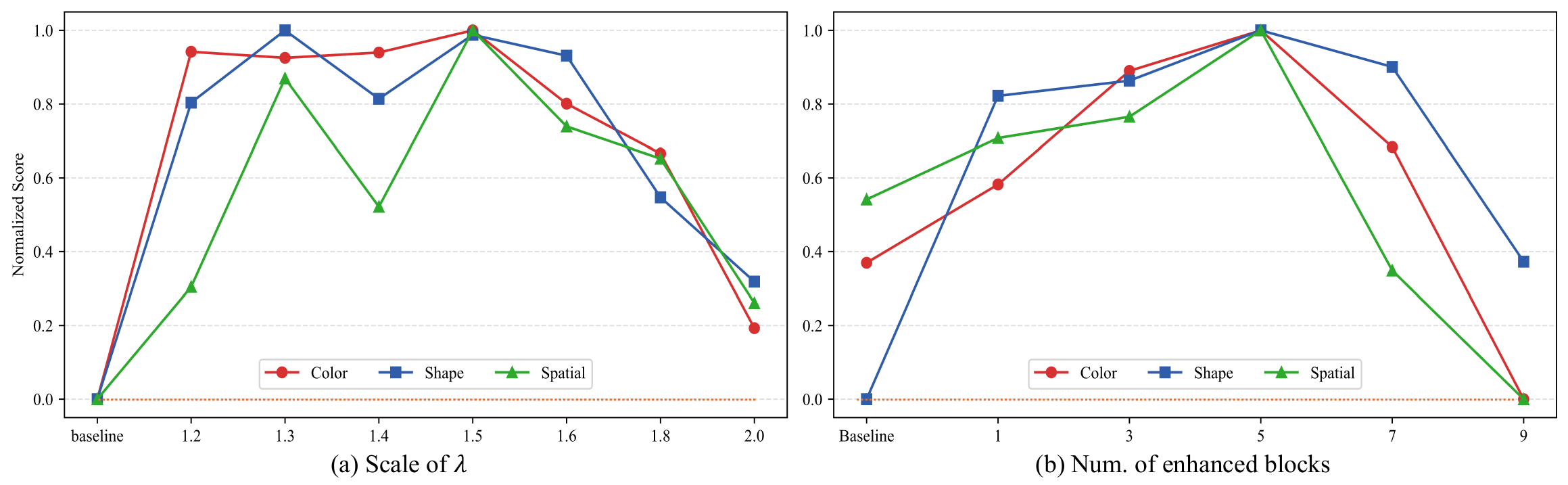} %
    \caption{
    \textbf{Ablation analysis on enhancing scale (a) and block selections (b)}. 
    %
    %
    }
    \label{fig:ablation_combined}
\end{figure}

\myparagraph{Analysis on $\lambda(l)$.}
This part investigates the sensitivity of the scale $\lambda(l)$ .
Specifically, we evaluate the performance of different attributes on FLUX with $\lambda(l)$ ranging from $1.2$ to $2.0$. 
As shown in Fig~\ref{fig:ablation_combined} (a), our method achieves significantly better results than the baseline despite some fluctuations, showing the effectiveness of our method.
Additionally, we also evaluate the performance of weakening the textual conditions in Tab.~\ref{tab:ablate_blocks_strength}.
It turns out that the weakening operation significantly decreases the model's performance, further demonstrating the importance of these vital blocks and validating the soundness of our method.

\myparagraph{Analysis on the selection of enhanced blocks.}
To evaluate the effectiveness of our analysis in selecting the proper number of blocks for enhancement, we apply our enhancement to varying block counts $N \in \{1,3,5,7,9\}$.
The results in Fig.~\ref{fig:ablation_combined} (b) show that increasing $N$ initially improves performance, yet manipulating more than $9$ blocks may degrade it due to distribution shift.
We further tailor random FLUX blocks ($5$ and $all$) instead of our selected ones, and the results are shown in Tab.~\ref{tab:ablate_blocks_strength}.
The table shows that enhancing random or all blocks performs worse than enhancing the dedicated blocks identified by our analysis, highlighting the effectiveness of our approach.
What's more, this observation also reflects that different blocks do not contribute equally to different attributes, consistent with our findings in Sec.~\ref{sec:analysis}.
%
%

\myparagraph{Robustness across blocks, samplers, timesteps and resolution.}
The tailoring blocks are identified once per model and remain fixed across all inference configurations.
Tab.~\ref{tab:robustness_inference} validates robustness across different samplers (EDM, dpm++), denoising steps ($\mathcal{S}$), and image resolutions ($\mathcal{R}$).
TexTailor yields consistent gains in all settings, confirming the robustness of block selection while inference hyperparameters change.

\begin{table}[t]
    \centering
    \scriptsize
    \caption{
    \textbf{T2I-CompBench Score across inference configurations on FLUX.}
    }
    \label{tab:robustness_inference}
    \setlength{\tabcolsep}{6pt}
    \begin{tabular}{l|cc|cccc|cc}
    \toprule
     & EDM & dpm++ & $\mathcal{S}$=8 & $\mathcal{S}$=15 & $\mathcal{S}$=28 & $\mathcal{S}$=50 & $\mathcal{R}$=512 & $\mathcal{R}$=1024 \\
    \midrule
    Baseline & 47.39 & 47.42 & 36.72 & 43.57 & 47.39 & 47.80 & 47.24 & 47.39 \\
    Ours     & \textbf{50.46} & \textbf{50.59} & \textbf{40.32} & \textbf{47.58} & \textbf{50.46} & \textbf{50.86} & \textbf{50.41} & \textbf{50.46} \\
    \bottomrule
    \end{tabular}
\end{table}

\begin{figure}[t]
    \centering
    \includegraphics[width=0.95\linewidth]{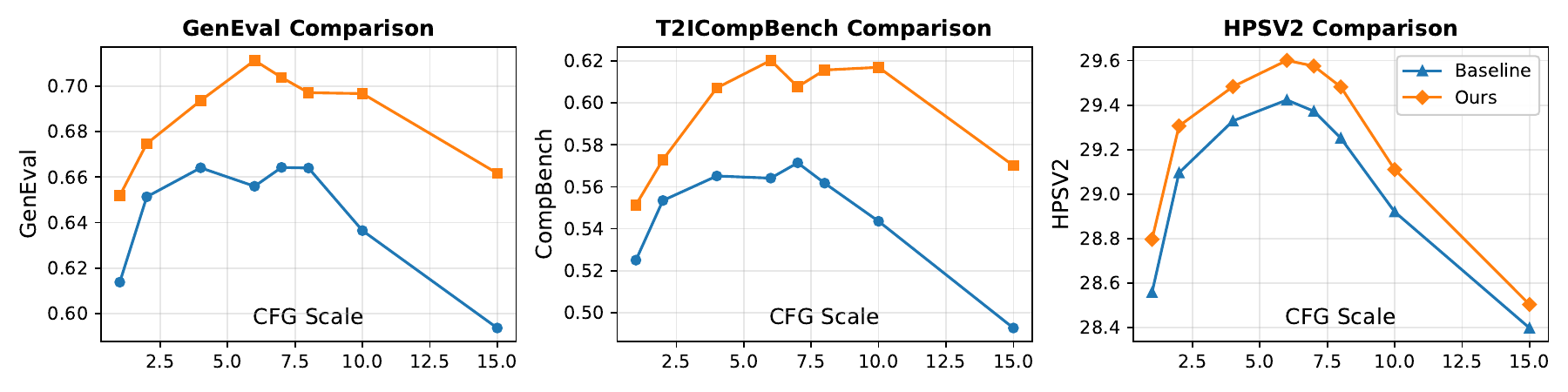}
    \caption{
    \textbf{Robustness to varying CFG values.}
    TexTailor consistently outperforms the FLUX baseline across all CFG settings on GenEval, T2I-CompBench, and HPSv2.
    }
    \label{fig:cfg_ablation}
\end{figure}

\myparagraph{Varying CFG Values.}
To verify that our improvements are not attributable to CFG upscaling effects, we evaluate our method on FLUX across a range of CFG values.
As shown in Fig.~\ref{fig:cfg_ablation}, our method consistently outperforms the baseline on GenEval, T2I-CompBench, and HPSv2 across all CFG settings, demonstrating that the gains originate from our targeted text hidden state enhancement rather than from guidance scaling.

\section{Applications}
\label{sec:applications}

\begin{figure}[t]
    \centering
    \includegraphics[width=0.99\textwidth]{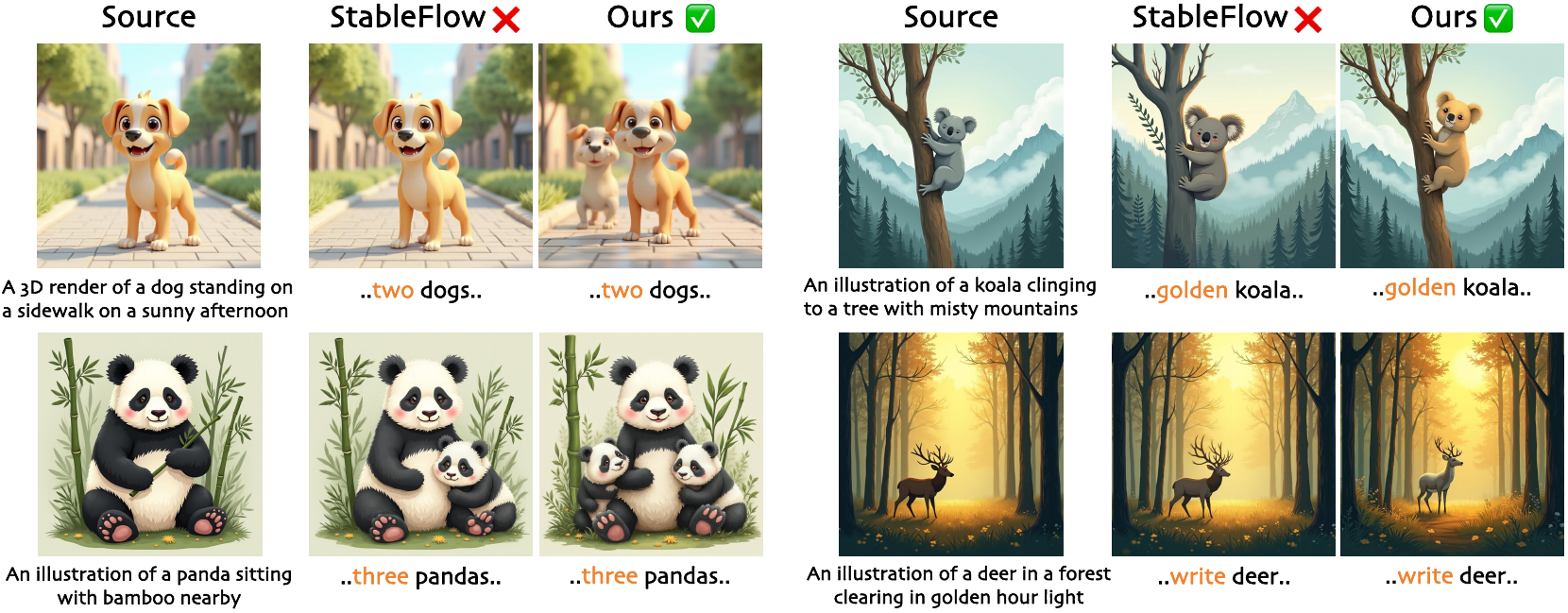} %
    \caption{
    \textbf{Qualitative comparisons of editing results between Stable Flow and our method}. 
    Our method enables more precise editing on specific attributes by changing the color, amount, \emph{etc.}
    }
    \label{fig:editing_qualitative}
\end{figure}

\myparagraph{Enabling Precise Text-based Editing.}
We incorporate our enhancement into editing tasks, facilitating precise textual editing with the target text instructions.
Following the Stable Flow~\cite{stableflow_avrahami2025}, we perform image editing via parallel generation, injecting self-attention features from the source image into the target image to preserve visual content.
Differently, our empirical findings motivate us to enhance target text embeddings across critical blocks to improve the editing accuracy.
The results in Fig.~\ref{fig:teaser}, Fig.~\ref{fig:editing_qualitative} and Tab.~\ref{tab:editing_result} show the effectiveness of our method.
Our method outperforms Stable Flow on CLIP$_{txt}$ score ($\uparrow$0.94), showing more accurate editing towards textural instructions.
Meanwhile, the CLIP$_{img}$ similarity remains nearly unchanged ($\downarrow$0.008), suggesting that our method effectively enables more precise editing in line with the given instructions while preserving the visual integrity and coherence of the images.
Furthermore, the result of human preference further reflects the effectiveness of our method.

\myparagraph{Inference Acceleration.}
Recall that our analysis indicates that removing some blocks causes a smaller impact on the output, suggesting their role in rendering fine-grained details instead of vital semantics.
Accordingly, we accelerate inference with a training-free mechanism by skipping specific blocks identified as less critical from our probing analysis.
Formally, we skip the corresponding block or its CFG operation during inference.
Notably, our method is complementary to existing acceleration techniques, such as Teacache~\cite{teacache_liu2024timestep}, to achieve further acceleration.
Tab.~\ref{tab:accelerating_time} reports inference acceleration results by skipping less critical blocks, showing averaged inference time over $400$ prompts on NVIDIA 4090 and H100 GPUs.
Our method substantially reduces inference time and can be seamlessly combined with TeaCache for further acceleration.
%
%
%
The minor drops of HPSv2 ($\downarrow 0.19$) and Aesthetic ($\downarrow 0.01$) scores are justified by the acceptable visual quality and faster inference time.

\begin{table}[t]
    \centering
    \scriptsize
    \begin{minipage}[c]{0.46\textwidth}
        \centering
        \caption{
            \textbf{Editing results.}
            }
            \label{tab:editing_result}
            \setlength{\tabcolsep}{1.5pt}
            \begin{tabular}{@{}lcc|c@{}}
            \toprule
            Method & CLIP$_{img}$ & CLIP$_{txt}$& Human Pref\\
            \midrule
            Stable Flow & \textbf{0.9642} & 35.2584 & 40.98\% \\
            + Ours      & 0.9637          & \textbf{36.1988} & \textbf{59.02\%} \\
            \bottomrule
            \end{tabular}%
    \end{minipage}
    \hfill
    \begin{minipage}[c]{0.50\textwidth}
        \centering
        \caption{\textbf{Acceleration results}.}
            \label{tab:accelerating_time}
            \setlength{\tabcolsep}{1.5pt}
            \begin{tabular}{lcc|cc}
            \toprule
            Method & T($s$,4090)& T($s$,H100)& HPSv2 & Aes. \\
            \midrule
            FLUX        & 36.79 & 13.09 & \textbf{29.0533} & \textbf{6.1903} \\
            + TeaCache    & 26.62 & 9.61  & 28.8951 & 6.2067 \\
            + Ours      & \textbf{24.53} & \textbf{8.88}  & 28.8647 & 6.1801 \\
            \bottomrule
            \end{tabular}%
    \end{minipage}
\end{table}

\section{Conclusions}

\label{sec:discussions}

%
%
In this work, we systematically analyze block-wise contributions and their interactions with text conditions, offering a better understanding of the internal mechanisms within MMDiT-based generative models.
Meanwhile, our analysis reveals several valuable findings that unlock new possibilities for improving the synthesis quality.
Based on these findings, we introduce novel training-free techniques to facilitate text alignment of T2I tasks, precise semantic manipulation of editing tasks, and accelerated inference in a training-free manner.
Extensive results demonstrate the effectiveness of our method.
%

%
Despite substantial performance gains, \method has limitations. It relies on automatic block-wise analysis and cannot perfectly synthesize highly complex prompts due to pretraining constraints. Future work could incorporate trainable modules and token-level dynamic routing to further improve performance.



\section*{Acknowledgements}

This work was supported in part by AI for Science Program, Shanghai Municipal Commission of Economy and Informatization (Grant No. 2025-GZL-RGZN-BTBX-02017) and the National Natural Science Foundation of China (Grant No.62576103).

%
%
\bibliographystyle{splncs04}
\bibliography{reference}

\clearpage
\appendix
\setcounter{page}{1}

\title{Supplementary Material for ``TexTailor: Inference-Time Textual Guidance Tailoring for Multimodal Diffusion Transformers''}
\maketitle

\renewcommand{\theHsection}{appendix.\Alph{section}}

\numberwithin{figure}{section}
\numberwithin{table}{section}
\numberwithin{equation}{section}

\section{Overview of Appendix}
The appendix provides supplementary information supporting the main text. We begin with a discussion on the limitations of our method (\cref{sec:limitations}). Next, we present a detailed analysis of probing results (\cref{appendix:probing_results_chapter}), including baseline models, implementation details, datasets, quantitative metrics such as amount and CLIP score, and evaluation procedures. The appendix (\cref{appendix:evaluation_details}) then details evaluation setups, covering inference configurations, benchmarks, token-level enhancement mask localization, editing tasks, and acceleration experiments. Additional quantitative results (\cref{appendix:more_quantitative_results}) include analyses on enhanced block selection, scaling schemes, ablation studies, and token-level enhancements. More qualitative results (\cref{appendix:more_qualitative_results}) show visual examples across models and tasks, including generation and editing. We also discuss observed failure cases and limitations. Finally, the human evaluation protocol for assessing output quality and alignment is described (\cref{appendix:human_evaluation_details}). Together, these sections provide a comprehensive resource for reproducing experiments, understanding model behavior, and exploring findings beyond the main text.

\section{Limitations}
\label{sec:limitations}
Although our method delivers substantial performance gains across multiple tasks, certain limitations remain.

\noindent \textbf{Dependence on the preliminary block-wise analysis.}
A core part of our enhancement pipeline relies on an initial, automatic analysis that identifies block-wise interactions inside the model. Luckily, our analysis is totally automatic and could be performed across various models with different numbers of blocks, varied model sizes, as well as different models (e.g., SD3.5, FLUX, Qwen Image).

\begin{figure}[htbp]
    \centering
    \vspace{-20pt}
    \setlength{\abovecaptionskip}{0pt}
    \setlength{\belowcaptionskip}{-10pt}
    \includegraphics[page=7,width=0.8\textwidth]{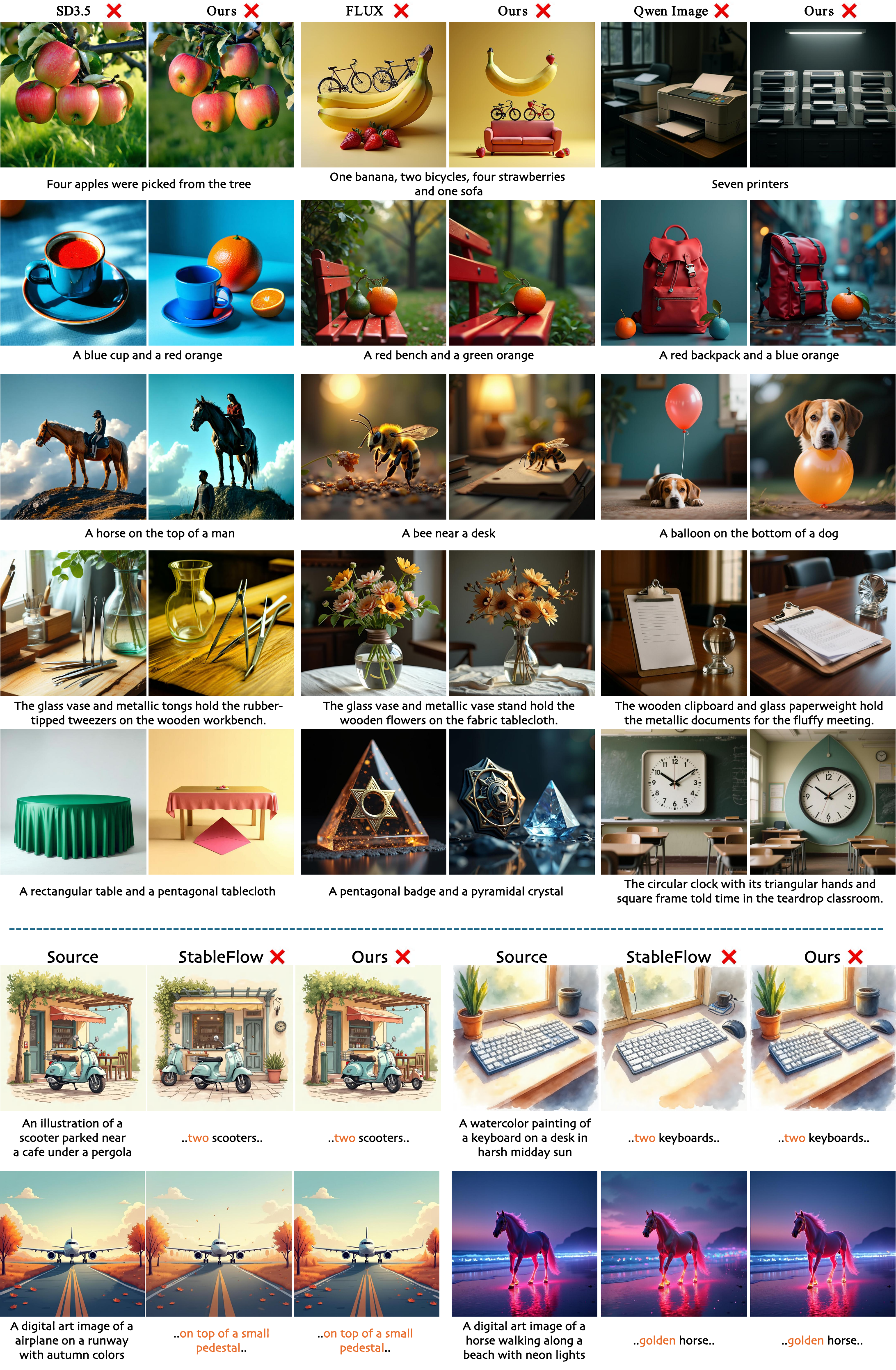}
    \caption{Failure cases of generation and editing. SD3.5 sometimes misidentifies attributes, confusing colors and objects. For rare real-world cases, our method may produce correct attributes but also hallucinations, e.g., a dog missing its body. In editing, hard cases mainly involve amount, reflecting the model's limited counting ability.}
    \label{fig:appendix_failure_cases}
    \vspace{-30pt}
\end{figure}

\noindent \textbf{Limited fidelity on very complex prompts.}
For extremely complex or highly detailed prompts, our method may fail to capture all fine-grained elements. This limitation mainly stems from the pretraining data distribution, where rare object combinations or subtle high-frequency details are underrepresented, sometimes resulting in missing elements or artifacts in the output.
As could also be observed from the failure cases in Fig.~\ref{fig:appendix_failure_cases}.

\noindent \textbf{Evaluation and generalization limits.}
Our evaluation on standard benchmarks (GenEval and T2IComBench++) may not fully reflect perceptual quality or robustness, and gains might not generalize to niche domains or prompts requiring world knowledge absent from pretraining.

\section{Detailed Analysis of Probing Results}
\label{appendix:probing_results_chapter}

\subsection{Baseline Models and Implementation Details}
\label{appendix:probing_implementation_details}
We implement our probing analysis based on the widely adopted MMDiT-based text-to-image models, including:
\begin{itemize}
    \item Stable Diffusion 3.5-Large
    ~\cite{SD35_2024}: A latent diffusion model with approximately 8 billion parameters, based on the Multimodal Diffusion Transformer (MMDiT) architecture. It demonstrates strong performance in prompt adherence, typography, and supports a mature ecosystem of extensions.
    \item FLUX.1-Dev
    ~\cite{flux_2024}: A 12B-parameter rectified-flow transformer model that adopts advanced training techniques and a substantially larger dataset to enhance visual fidelity. It has attracted significant community attention for its improvements in prompt alignment, detail rendering, and efficient sampling.
    \item Qwen Image
    ~\cite{qwenimage_wu2025technicalreport}: A 20B-parameter MMDiT model developed within the Qwen series, designed for robust multimodal reasoning and high-quality image synthesis. It is particularly noted for its strong performance in complex text rendering (especially Chinese) and text-guided image editing.
\end{itemize}

We use the official checkpoints provided by the authors and the \textit{diffusers} library for implementation. During inference, model weights are loaded in 16-bit precision. No acceleration techniques such as xformers or memory-efficient attention are used. The default parameters during inference are summarized in Tab. \ref{tab:appendix_default_parameters}.

\begin{table}[htbp]
    \centering
    \caption{Model information and default parameters during inference.}
    \label{tab:appendix_default_parameters}
    \setlength{\tabcolsep}{4pt}
    \begin{tabular}{lccc}
    \toprule
    Models          & SD3.5-large & FLUX.1-Dev  & Qwen Image \\
    \midrule
    MMDiT Blocks    & [0,37]  & [0,57]   & [0,59]      \\
    Parameters      & 8B          & 12B         & 20B        \\
    Inference Steps  & 28          & 28          & 50         \\
    CFG Scale    & 7.0         & 3.5         & 4.0        \\
    Size            & (1024,1024) & (1024,1024) & (1024,1024) \\
    \bottomrule
    \end{tabular}%
\end{table}

During probing analysis, we use identical hyperparameters for all models to ensure fair comparison, as summarized in Tab. \ref{tab:appendix_default_parameters}. For each model with $N$ blocks, we fix a random seed and generate one baseline image, $N$ \textit{disable} images, $N$ \textit{remove} images, and $N$ \textit{enhance} images—constituting one experimental group. For each of the three MMDiT-based models, we conduct five experimental groups using five different random seeds $(0, 42, 329, 1234, 99514)$. Final results are reported as the average across these five groups.

\subsection{Constructed Dataset for Probing Analysis}
\label{appendix:dataset_construction}
We construct a prompt dataset from the widely adopted HPDv2~\cite{HPSV2_wu2023human} and COCO~\cite{lin2014microsoftcoco} datasets, comprising $633$ diverse and complex prompts across three attributes: color, spatial relationship, and amount. For color attributes, we focus on objects with distinctive colors (e.g., ``red apple'', ``yellow banana''). For spatial relationships, we include prompts describing eight positional relations (e.g., ``left'', ``right'', ``above'', ``below'', ``upper left'', ``upper right'', ``lower left'', ``lower right''). For shape attributes, we cover a range of shapes (e.g., ``circle'',``oval'',``square'',``rectangle''). We also ensure diversity in object categories, including human, animal, natural scenes, indoor scenes, food, clothing \& accessories, vehicles, and so on. 

\begin{figure}[htbp]
    \centering
    \begin{subfigure}{0.42\textwidth}
        \centering
        \includegraphics[page=1,width=\linewidth]{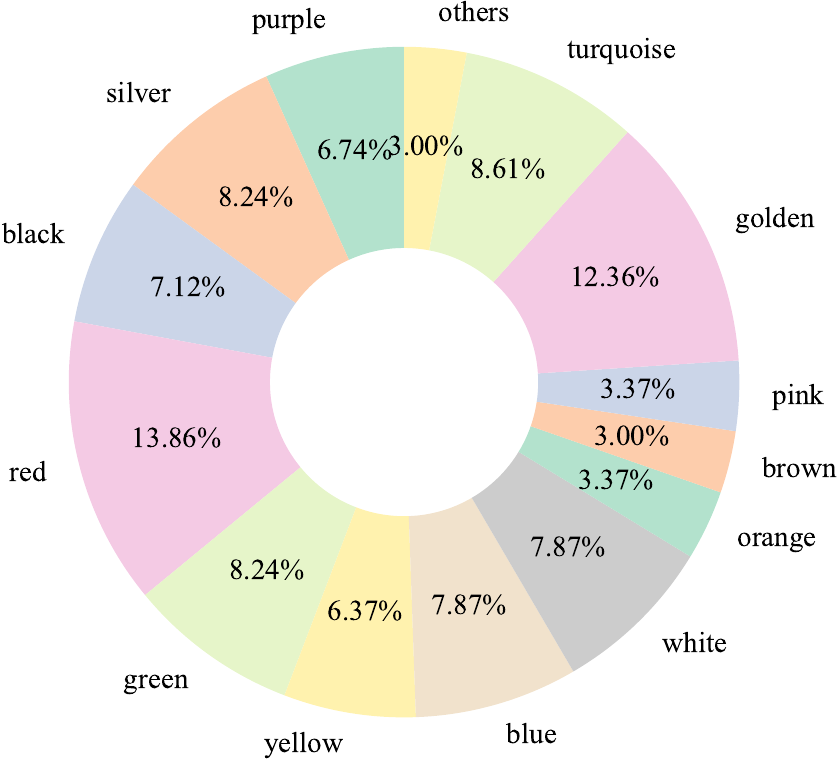}
        \caption{Color distribution.}
        \label{fig:appendix_color_piechart}
    \end{subfigure}
    \hfill
    \begin{subfigure}{0.5\textwidth}
        \centering
        \includegraphics[page=2,width=\linewidth]{figures/appendix_prompts_piecharts.pdf}
        \caption{Spatial distribution.}
        \label{fig:appendix_spatial_piechart}
    \end{subfigure}

    \begin{subfigure}{0.45\textwidth}
        \centering
        \includegraphics[width=\linewidth]{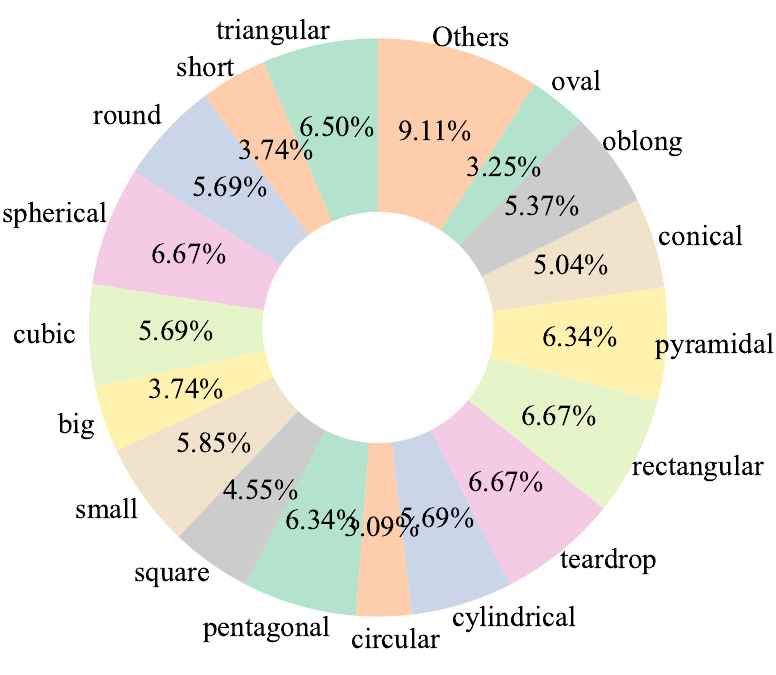}
        \caption{Shape distribution.}
        \label{fig:appendix_shape_piechart}
    \end{subfigure}
    \hfill
    \begin{subfigure}{0.48\textwidth}
        \centering
        \includegraphics[page=4,width=\linewidth]{figures/appendix_prompts_piecharts.pdf}
        \caption{Text object categories.}
        \label{fig:appendix_object_category}
    \end{subfigure}
    \caption{Statistics of our constructed datasets for probing analysis. Each subfigure presents the distribution of prompts for a specific attribute: color, spatial relation, shape, and object category.}
    \label{fig:appendix_prompts}
\end{figure}

The distribution of prompts across these attributes is illustrated in Fig. \ref{fig:appendix_prompts}. We ensure that the prompts are diverse and challenging, covering various object categories, colors, spatial relations, and quantities.

\subsection{Probing Analysis Results of Qwen Image}
\label{appendix:probing_more_results}
\begin{figure}[h]
    \centering

    \includegraphics[page=3, width=0.8\linewidth]{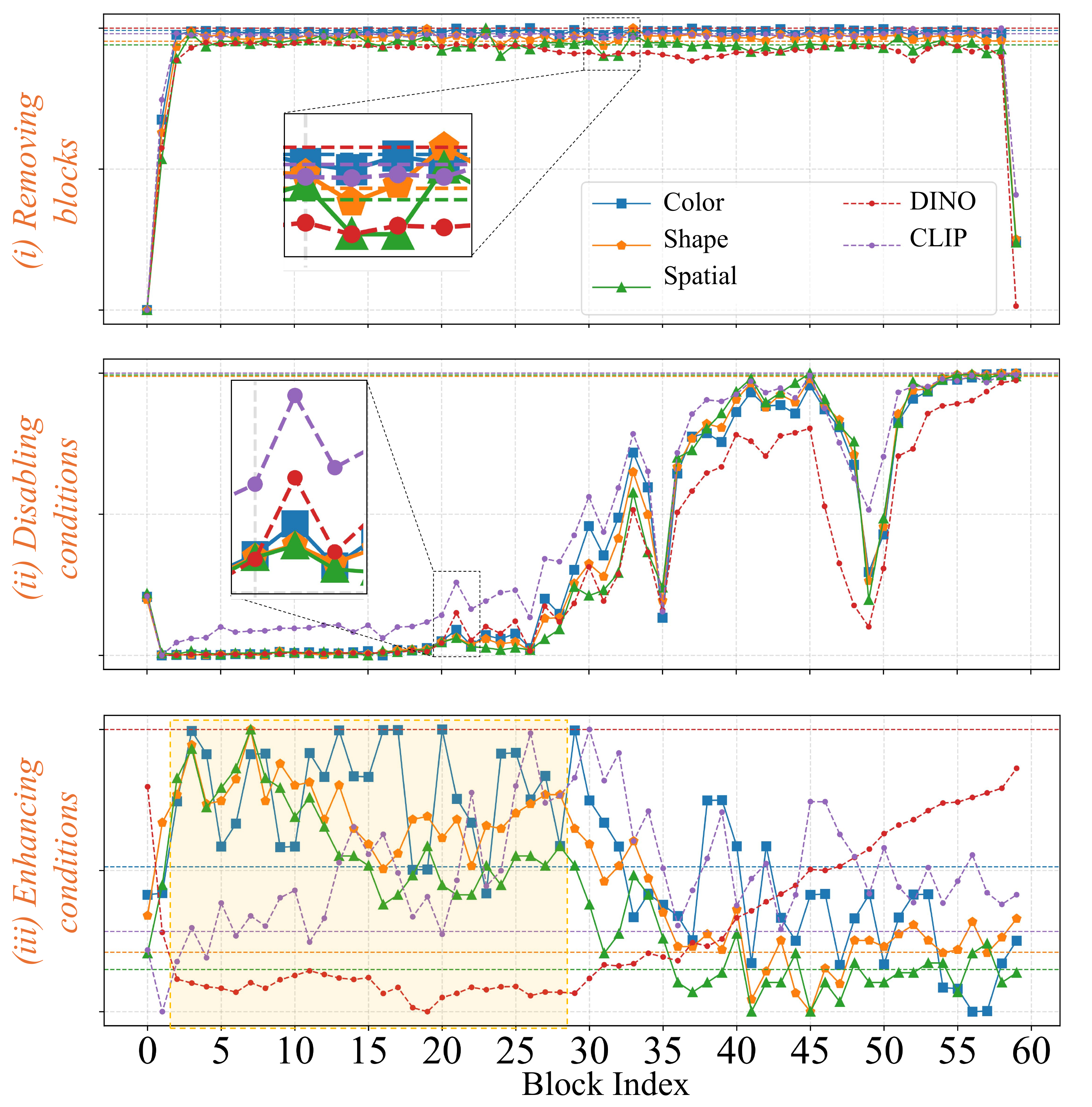}

    \caption{
        \textbf{Probing analysis of Qwen Image.}
        From top to bottom, the probing strategies are \textit{removing}, \textit{disabling}, and \textit{enhancing}.
        We choose three attributes: ``color'', ``shape'', and ``spatial relationship'' as representatives. 
        The results are averaged over five runs with different random seeds.
        The accuracy of different attributes is evaluated by QwenVL-2.5-72B on multiple runs, and DINOv2 and CLIP score shows the perceptual similarities.
        Enhancing in the yellow region leads to significant improvement on three attributes. The results are consistent with the main text.
        }
    \label{fig:appendix_probing_more_results}
\end{figure}

Due to space limitations, the main paper only presents the block-wise analysis results for SD3.5-Large and FLUX.1-Dev models. Here, we provide the additional results for Qwen Image model in Fig.~\ref{fig:appendix_probing_more_results}.
Notably, the Qwen Image model exhibits similar block-wise analysis results as the SD3.5-Large and FLUX.1-Dev models, indicating that the probing analysis is effective and generalizable across different models.
However, the \textit{removing} and \textit{enhancing} strategies show a marginal difference compared to the baseline model, due to the stronger fundamental capacity of Qwen Image.
The overall CLIP score trends are consistent with those of DINOv2 similarity, further validating the effectiveness and stability of our method.

\subsection{Details of Evaluation for Probing Analysis}

In probing analysis, we use the open-sourced Qwen2.5-VL 72B
~\cite{qwen25VL_bai2025} model for ``color'', ``shape'', and ``spatial relationship'' evaluation. 
We design specific systematic prompts to guide the model in accurately assessing whether the generated images align with the intended attributes in the text prompts. 
The shape evaluation prompt is similar to the color evaluation prompt.
The detailed prompts for color, and spatial relationship evaluation are provided in following colored boxes.

\begin{tcolorbox}[
    width=0.9\linewidth,
    center,
    colback=black!5!white,
    colframe=black!75!black,
    title=Color Evaluation Prompt
    ]
    \scriptsize
You are given an image, its caption, and a set of objects with their expected colors.

Your task: 1).For each object: check if the color in the caption matches the actual color in the image. 2).If the color matches, return "Yes". If the color does not match or the object is not visible, return "No".

Rules:
\begin{itemize}
    \item Output ONLY a single valid JSON object.
    \item The JSON keys must be exactly the provided object names.
    \item The values must be strictly "Yes" or "No".
    \item Do not generate any other words.
    \item Do not add explanations, extra text, or formatting outside the JSON.
\end{itemize}

Example:
\begin{center}
    \includegraphics[page=1,width=0.8\linewidth]{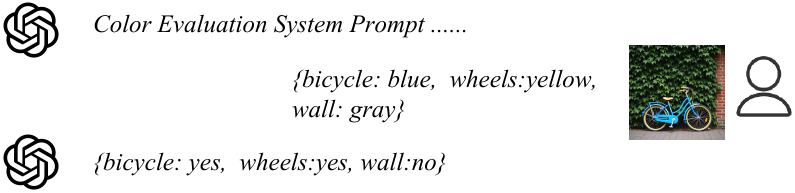}
\end{center}

\end{tcolorbox}

\begin{tcolorbox}[
    width=0.9\linewidth,
    center,
    colback=black!5!white,
    colframe=black!75!black,
    title=Spatial Evaluation Prompt
    ]
    \scriptsize

You are given an image, its caption, and a question about the spatial relationship between two objects in the image.

Your task: 1).Check whether the spatial relationship described in the question can be confirmed from the image.2).If the relationship is clearly visible and correct, return "Yes".3).If the relationship is not correct, cannot be seen, or the objects are unclear, return "No".

Rules:
\begin{itemize}
    \item Output ONLY a single string. The value must be strictly "Yes" or "No".
    \item Do not generate any other words.
    \item Do not add explanations, extra text, or formatting outside the answer.
\end{itemize}

Example:
\begin{center}
    \includegraphics[page=2,width=0.8\linewidth]{figures/appendix_human_interface.pdf}
\end{center}

\label{appendix:color_spatial_evaluation_prompt}
\end{tcolorbox}

\paragraph{Statistical Significance.}
We acknowledge that both the LLM-based evaluation have inherent limitations. LLMs may misinterpret visual details or be affected by biases in their training data. To address these issues, we conduct multiple experimental runs with different random seeds and report averaged results, thereby reducing the impact of individual evaluation errors. 
To further ensure the validity and robustness of our conclusions, we additionally employ alternative evaluation methods that are independent of the primary approaches. This cross-validation helps to mitigate the influence of dataset bias and evaluation inaccuracies on our experimental findings.

\section{Detailed Setup of Evaluation Results}
\label{appendix:evaluation_details}
\subsection{Inference Details}

During evaluation of text-to-image generation, editing, and acceleration, we use the same hyperparameters in Tab. \ref{tab:appendix_default_parameters}. The enhancement strength $\lambda$ is set to 1.5 by default. For the enhanced blocks, we select the block index in Tab. \ref{tab:enhance_blocks}. 
For full sentence tailoring, we set the mask $M$ to be a full one vector, which has already demonstrated strong performance.
For token-level tailoring, we construct the enhancement mask $M$ by passing the target phrases (\emph{e.g.}, ``two apple'',``red apple'', ``rectangle clock'').

\subsection{Evaluation Benchmarks}

For text-to-image generation, we evaluate our method on the widely used T2I-CompBench++~\cite{t2i_compbench_plusplus_huang2025t2i} and GenEval~\cite{geneval_ghosh2023} benchmarks. T2I-CompBench++ contains $8,000$ compositional prompts spanning color, spatial, 3D spatial, shape, texture, non-spatial relations, numeracy, and complex attributes. It extends the original benchmark~\cite{t2i_compbench_huang2023t2i} and introduces more challenging tasks (e.g., 3D spatial and numerical compositionality). We use all prompts and generate one sample per prompt. GenEval consists of $553$ structured prompts targeting single-object, two-object, counting, color, position, and color-attribute binding. Each prompt is paired with four generated samples, and performance is computed with an automatic evaluation pipeline based on object detection, counting, and attribute classification, providing interpretable error types (e.g., missing objects, incorrect color, or mis-counting).

We evaluate image quality and text-to-image alignment using LAION Aesthetic v2~\cite{laion_aesthetic_2022} and HPSv2~\cite{HPSV2_wu2023human}. LAION Aesthetic v2 measures visual appeal, while HPSv2 evaluates prompt-image alignment relative to human judgments. As these metrics capture different aspects, we report both and supplement them with task-specific evaluations and human studies to ensure a comprehensive assessment.

\subsection{Token-level Enhancement Mask Localization}
\label{appendix:token_level_mask}

\begin{figure}[htbp]
    \centering
    \includegraphics[page=1,width=0.95\textwidth]{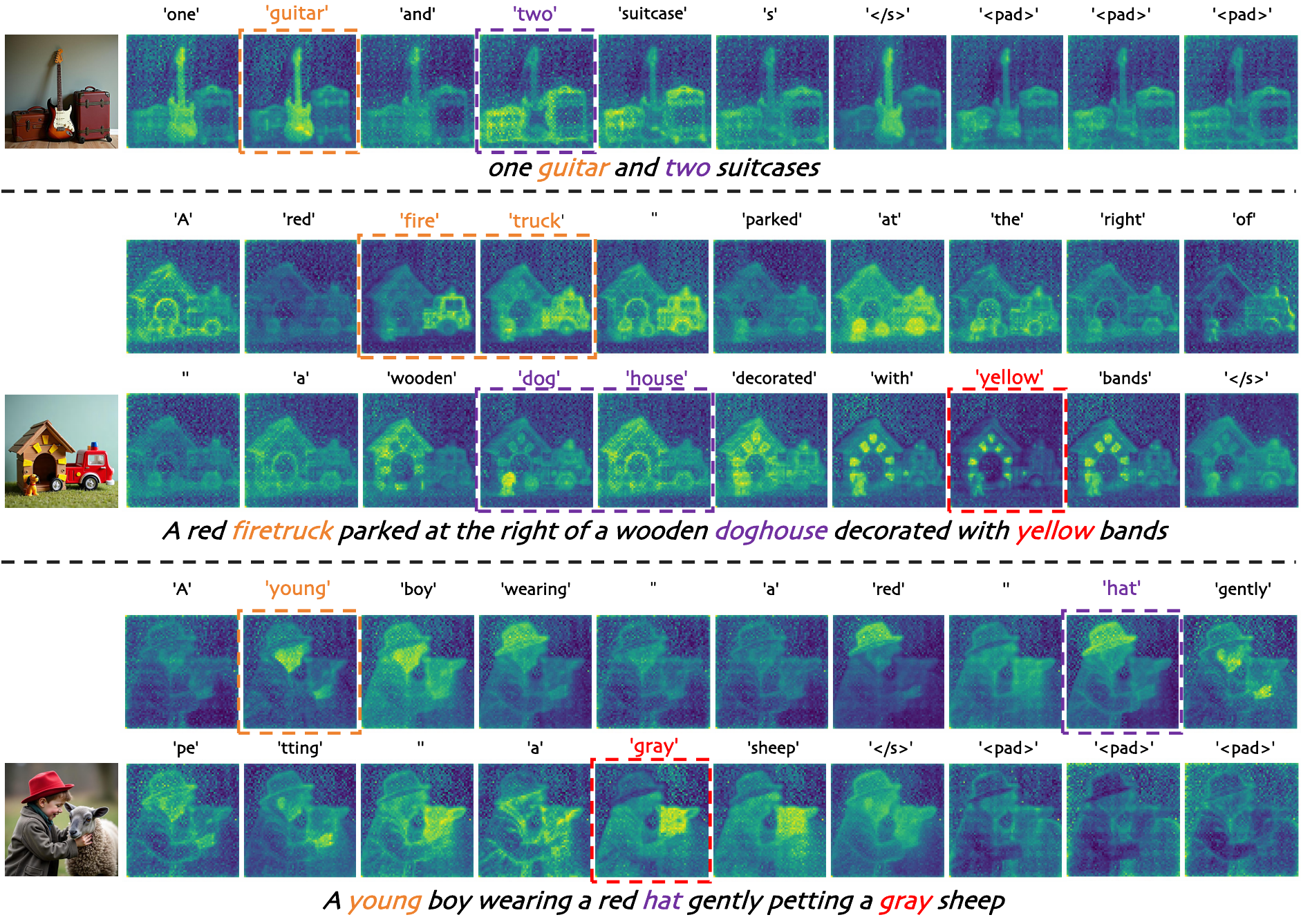}
    \caption{Visualization of token-level attention maps. Colored boxes indicate the target tokens in the prompt. The attention maps highlight the corresponding text tokens' indices in the image.}
    \label{fig:appendix_attention_maps}
\end{figure}

During inference, we record the multi-head self-attention of the concatenated features $Z_{in}$ at each MMDiT block and denoising step. To obtain a stable token-region corresponding, we aggregate the attention maps across all heads and denoising steps. Eventually, we get attention maps of shape $[N, H\times W, T]$, where $N$ is the number of MMDiT blocks, $H\times W$ is the spatial dimension of the image features, and $T$ is the number of text tokens. For visualization, we average all the MMDiT blocks' attention maps to get a single attention map of shape $[H\times W, T]$. We then normalize the attention maps along the spatial dimension and then resize them to the original image size. The visualization results are shown in Fig.\ref{fig:appendix_attention_maps}. We can see that 'dog' and 'cat' tokens have high attention values in the corresponding image regions, indicating that the token-level enhancement can effectively target specific areas in the image.

Based on the above analysis, we first tokenize the input prompt 
$\mathcal{P}$ using the same tokenizer as the MMDiT model, obtaining a sequence of token IDs $\mathbf{P} = [p_1, p_2, \dots, p_N]$. Given a target phrase $\mathcal{Q}$ (e.g., ``brown'', ``firetruck''), we tokenize it as $\mathbf{Q} = [q_1, q_2, \dots, q_M]$. We then search for all subsequences in $\mathbf{P}$ that match $\mathbf{Q}$. The starting indices of these matches are collected in the set $\mathcal{I} = \{ i \mid (p_i, p_{i+1}, \ldots, p_{i+M-1}) = (q_1, q_2, \ldots, q_M) \}$. The mask $M$ is constructed with the following rule:
\begin{equation*}
M_j =
\begin{cases}
1, & \text{if } \mathcal{I} = \varnothing \text{ or } \exists (i,i+M-1) \in \mathcal{I} \\
\qquad &\qquad  \text{ with } j \in [i,i+M-1],\\ 
0, & \text{otherwise}.
\end{cases}
\end{equation*}

Here, $M_j$ indicates whether the $j$-th token in the prompt should be enhanced. If not matched, the token-level enhancement defaults to sentence-level enhancement by setting all entries of $M$ to 1.

\subsection{Editing details}
Stable Flow~\cite{stableflow_avrahami2025} is adopted as the baseline, which selects vital blocks based on the perceptual similarity. 
However, its block selection is not task-specific and may be inaccurate for fine-grained editing, whereas our approach leverages probing analysis to identify blocks tailored to the editing task.
Prior benchmarks lack coverage of quantity, attribute binding, and spatial relationship editing. 
To address this, we construct a new editing benchmark comprising $1,000$ images and corresponding editing prompts. 
Each source prompt is paired with four target prompts, covering object addition (with varying colors), background changes, color and lighting adjustments, shape and direction modifications, %
positional changes, quantity variations, and object actions. 
Evaluation is performed for each image-prompt pair using CLIP-image similarity and CLIP score to assess image quality and prompt adherence. Human evaluation details are provided in Sec.\ref{appendix:human_evaluation_details}.

\subsection{Acceleration Details}

Our method is conceptually related to TeaCache~\cite{teacache_liu2024timestep}, which accelerates video diffusion by using timestep embeddings to estimate output differences and cache intermediate results selectively. In contrast, we skip blocks deemed irrelevant for the current editing task based on probing analysis. While TeaCache reduces redundancy across timesteps, our approach reduces computation across feature blocks, enabling acceleration without affecting editing quality.

\begin{figure}[htbp]
    \centering
    \includegraphics[page=2,width=\textwidth]{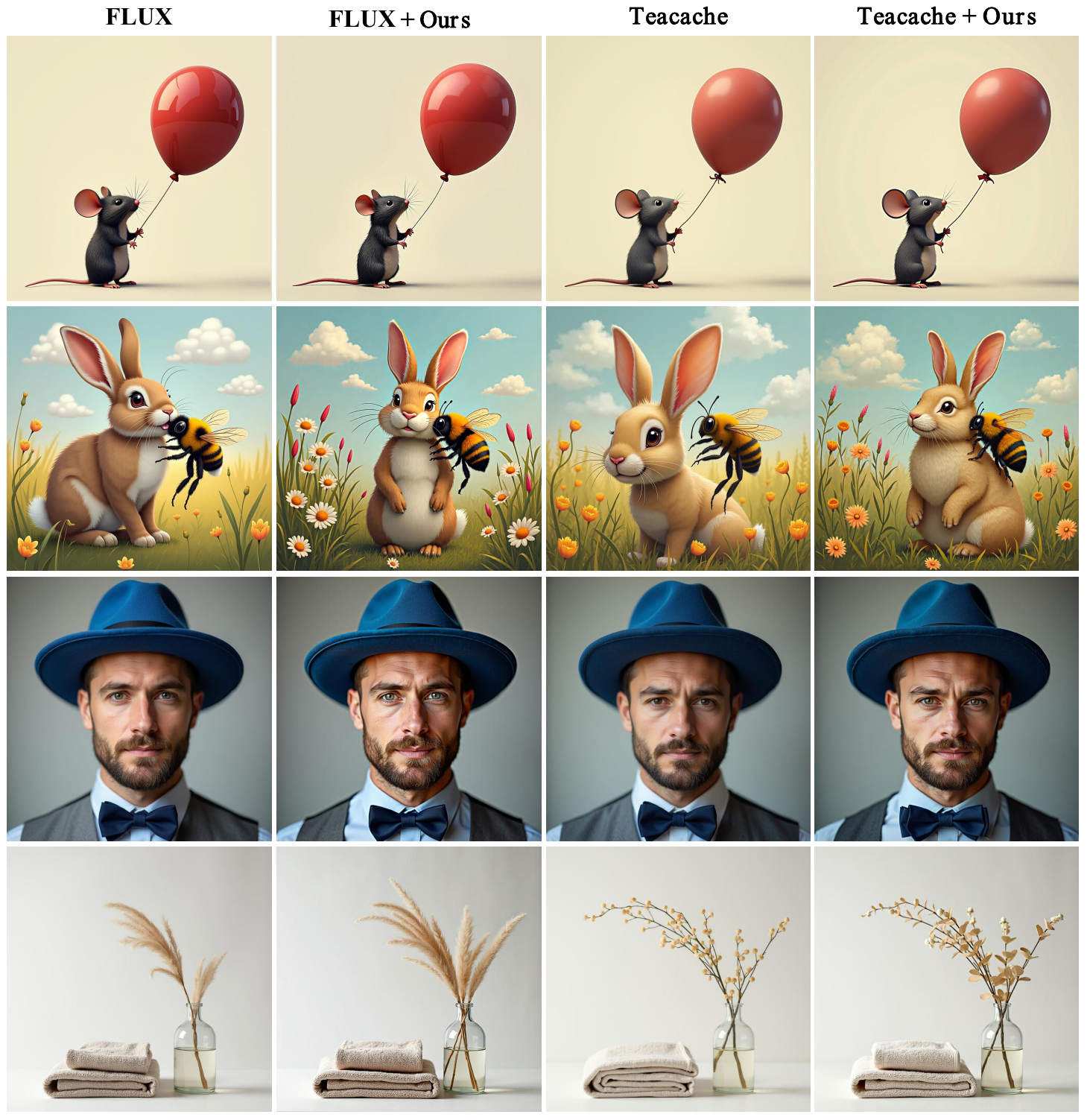}
    \caption{Examples of accelerated generation. 
    Our method removes certain blocks, which may lead to different details in the generated image, yet the overall synthesis quality and textual semantics are consistent with the user prompts.
    The four prompts are (1) ``a balloon on the right of a mouse'', (2) ``a rabbit hidden by a bee'', (3) ``a man in a blue and blur hat with a gray shirt and bowtie'', (4) ``a fabric towel and a glass vase''.}
    \label{fig:appendix_acceleration_examples}
\end{figure}

For acceleration, we skip one-third of the CFG steps and remove three MMDiT blocks. Applying our method to both the FLUX baseline and TeaCache demonstrates significant speedup while maintaining comparable image quality. In each experiment, we randomly select $400$ prompts from T2I-CompBench++ and generate one sample per prompt. The CFG steps skipped are $[5, 10, 15, 20, 25, 30, 35, 40, 45, 50]$, and the removed MMDiT blocks are $[30, 40, 50]$. We repeat the experiments on both NVIDIA 4090 and H100 GPUs to verify the stability and robustness of our approach. We present additional examples of accelerated generation in Fig. \ref{fig:appendix_acceleration_examples}. The results in Tab.~\ref{tab:accelerating_time} and Fig.~\ref{fig:appendix_acceleration_examples} demonstrate that our method can seamlessly integrate with TeaCache, achieving significant speedup while maintaining high-quality generation.

\section{More Quantitative Results}
\label{appendix:more_quantitative_results}

\subsection{Original Data for Enhancing Scale and Block Ablation}

In Fig.~\ref{fig:ablation_combined}, we present experiments conducted with a fixed enhancing scale ranging from $1.2$ to $2.0$, as well as experiments varying the number of enhancing blocks. In the main text, the vertical axes were normalized to emphasize consistent trends across different models and evaluation metrics. Here, we provide the original, unnormalized data for reference (see Tab.~\ref{tab:appendix_enhancing_original} and Tab.~\ref{tab:appendix_number_block_original} for details). In addition, we include the corresponding results for SD3.5 under the same enhancing scale settings.
%
%
Increasing the enhancing scale improves performance up to a point, after which gains plateau or slightly decline, showing a trade-off between strength and stability.

\begin{table}[htbp]
    \centering
    \small
    \setlength{\tabcolsep}{4pt}
    \caption{Results of the enhancing scale experiments. Except the FLUX data presented in the main text, we also provide the corresponding results for SD3.5 under the same experimental settings.}
    \label{tab:appendix_enhancing_original}
    \begin{tabular}{l|lcccccccc}
    \toprule
    \multicolumn{2}{c}{Methods/$\lambda(l)$} &
      Baseline & 1.2 & 1.3 & 1.4 & 1.5 & 1.6 & 1.8 & 2.0 \\
    \midrule
    \multirow{3}{*}{FLUX} & color   & 0.7322 & 0.7776 & 0.7768 & 0.7775 & \textbf{0.7804} & 0.7708 & 0.7643 & 0.7415 \\
      & shape   & 0.4908 & 0.5389 & 0.5489 & 0.5381 & \textbf{0.5482} & 0.5449 & 0.5226 & 0.5093 \\
      & spatial & 0.6603 & 0.6827 & 0.7244 & 0.6987 & \textbf{0.7340} & 0.7147 & 0.7083 & 0.6795 \\
    \midrule
    \multirow{3}{*}{SD3.5} & color & 0.7284 & 0.7992 & 0.8064 & \textbf{0.8072} & 0.8052 & 0.7874 & 0.7785 & 0.7608 \\
      & shape   & 0.5592 & 0.6432 & 0.6656 & 0.6642 & \textbf{0.6744} & 0.6659 & 0.6255 & 0.6048 \\
      & spatial & 0.6418 & 0.6683 & 0.7596 & 0.7716 & 0.7885 & \textbf{0.7933} & 0.7740 & 0.7486 \\
    \bottomrule
    \end{tabular}
\end{table}

\begin{table}[htbp]
    \centering
    \small
    \setlength{\tabcolsep}{4pt}
    \caption{Results of experiments varying the number of enhancing blocks on FLUX.}
    \label{tab:appendix_number_block_original}
    \begin{tabular}{lcccccc}
    \toprule
    Methods & Baseline & 1 & 3 & 5 & 7 & 9\\
    \midrule
    color   & 0.7365 & 0.7533 & 0.7776 & \textbf{0.7863} & 0.7613 & 0.7073 \\
    shape   & 0.4720 & 0.5159 & 0.5181 & \textbf{0.5254} & 0.5201 & 0.4919 \\
    spatial & 0.2985 & 0.3109 & 0.3152 & \textbf{0.3327} & 0.2841 & 0.2581 \\
    \bottomrule
    \end{tabular}
\end{table}

\subsection{More Results about Token-level Enhancement}

As shown in Table~\ref{tab:appendix_tokenlevel}, token-level enhancement generally provides more precise guidance compared to sentence-level enhancement, leading to consistent but modest improvements across amount-related attributes. The gains, however, remain limited, which can be attributed to the intrinsic weakness of current diffusion models in numerical reasoning and counting. This suggests that while finer-grained control enhances alignment with quantitative instructions, addressing the fundamental limitation of numerical competence in the base models remains an open challenge.

\begin{table}[htbp]
    \centering
    \small
    \caption{Token-level vs. sentence-level enhancement on amount-related attributes.}
    \label{tab:appendix_tokenlevel}
    \setlength{\tabcolsep}{4pt}
    \begin{tabular}{lcc}
    \toprule
    Methods        & Amount(T2I-CompBench++) & Count(GenEval)  \\
    \midrule
    SD3.5          & 0.5987 & 0.6344 \\
    ours(sentence) & 0.5929 & 0.6125 \\
    ours(token)    & \textbf{0.6088} & \textbf{0.6375} \\
    \midrule
    FLUX           & 0.5877 & 0.6375 \\
    ours(sentence) & 0.5860 & 0.6000 \\
    ours(token)    & \textbf{0.6091} & \textbf{0.6438} \\
    \midrule
    Qwen Image     & 0.7406 & 0.8562 \\
    ours(sentence) & 0.7359 & 0.8275 \\
    ours(token)    & \textbf{0.7616} & \textbf{0.8594} \\
    \bottomrule
    \end{tabular}%
\end{table}

\subsection{Detailed Acceleration Results}
In the main paper Tab. \ref{tab:accelerating_time}, we evaluate the results of our method for inference acceleration by removing less critical blocks. Here we provide more detailed results in Tab. \ref{tab:appendix_acceleration}, including the time cost on both NVIDIA 4090 and H100 GPUs, and the image quality metrics (HPSV2, LaionAesthetic V2, CLIP-Text) on different models. We test two baseline models: the original FLUX.1-Dev and the TeaCache-optimized version. For each baseline, we apply our method with different CFG step skipping and block removal strategies.

\begin{table}[htbp]
    \centering
    \small
    \caption{Details of acceleration }
    \label{tab:appendix_acceleration}
    \setlength{\tabcolsep}{4pt}
    \resizebox{\linewidth}{!}{%
    \begin{tabular}{lccccc}
    \toprule
    Method & Time(4090)$\downarrow$ & Time(H100)$\downarrow$ & HPSV2$\uparrow$  & LaionAes V2$\uparrow$ & CLIP-Text$\uparrow$ \\
    \midrule
    FLUX  & 36.7889 & 13.0876 & \textbf{29.0533}  & 6.1903 & \textbf{26.9986} \\
    skip CFG $seq(5,50,10)$ & 35.5859 & 12.6414 & 29.0395  & \textbf{6.2081} & 26.9761 \\
    skip CFG $seq(5,55,5)$  & 33.6433 & 11.9846 & 28.7259  & 6.1340 & 26.8841 \\
    skip CFG $seq(6,58,3)$ & \textbf{31.6931} & \textbf{11.3010} & 28.8408  & 6.1034 & 26.7460 \\
    Ours & 33.3387 & 11.8734 & 28.9212  & 6.1874 & 26.7807 \\
    \midrule
    Teacache                              & 26.6187 & 9.6125  & 28.8951 & 6.2067 & 26.8346 \\
    skip CFG $seq(5,50,10)$               & 26.1385 & 9.4414  & 28.9054 & 6.1983 & 26.8646 \\
    skip CFG $seq(5,55,5)$               & 25.2994 & 9.1565  & 28.8822 & 6.1973 & \textbf{26.8984} \\
    skip CFG $seq(6,58,3)$  & \textbf{24.5276} & \textbf{8.8804}  & 28.8647  & 6.1801 & 26.7946 \\
    Ours & 24.9992 & 9.0743  & \textbf{28.9481} & \textbf{6.2256} & 26.7722 \\
    \bottomrule
    \end{tabular}%
    }
\end{table}

The CFG steps skipped are $seq(5,50,10)$, $seq(5,55,5)$, and $seq(6, 58, 3)$, and the removed MMDiT blocks are $[30, 40, 50]$, where $seq(a,b,c)$ denotes the arithmetic sequence starting from $a$ to $b$ with step $c$. We can see that skipping more CFG steps leads to faster inference without significantly affecting image quality. We finally choose to skip CFG $seq(6, 58, 3)$ steps and remove $[30, 40, 50]$ blocks as the default setting for a good trade-off between speed and quality.


\subsection{Analysis on the selection of enhanced blocks}
In the main paper, Tab. \ref{tab:ablate_blocks_strength} only shows the results of color, shape, and spatial of different enhanced blocks and different strengthening scales. Here we provide the full results of T2I-CompBench++ in Tab. \ref{tab:appendix_enhance_layers_selection}. 

In the weakening experiments, using the same enhanced blocks as Tab.~\ref{tab:enhance_blocks}, all metrics decline relative to the baseline and $0.7$/$0.9$ scales, confirming the importance of these blocks for compositional generation. Randomly strengthening five blocks yields some improvement, but is less effective than our selected blocks, supporting the effectiveness of our probing approach. Strengthening all blocks significantly reduces performance, likely due to over-enhancement disrupting feature distribution.

\begin{table}[htbp]
    \centering
    \caption{Full results of small $\lambda(l)$ enhancement and block selection experiments.  All the experiments are conducted with FLUX on T2I-CompBench++.}
    \resizebox{\linewidth}{!}{%
    \begin{tabular}{@{}ccccccccc@{}}
    \toprule
    \multicolumn{1}{c}{\multirow{2}{*}{\textbf{Methods}}} &
    \multicolumn{3}{c}{\textbf{Attribute Binding}} &
    \multicolumn{3}{c}{\textbf{Object Relationship}} &
    \multicolumn{1}{c}{\multirow{2}{*}{\textbf{Amount}}} &
    \multicolumn{1}{c}{\multirow{2}{*}{\textbf{Complex}}} \\ 
    \cmidrule(lr){2-4} \cmidrule(l){5-7}
    \multicolumn{1}{c}{} &
    \multicolumn{1}{c}{Color} &
    \multicolumn{1}{c}{Shape} &
    \multicolumn{1}{c}{Texture} &
    \multicolumn{1}{c}{2D Spatial} &
    \multicolumn{1}{c}{3D Spatial} &
    \multicolumn{1}{c}{Non-Spatial} &
    \multicolumn{1}{c}{} &
    \multicolumn{1}{c}{} \\
    \midrule
    FLUX     & 0.7322 & 0.4908 & 0.6490 & 0.2935 & 0.3739 & 0.3044 & \textbf{0.5877}  & 0.3597 \\
    0.7 & 0.3161 & 0.2653 & 0.2707 & 0.1100 & 0.1831 & 0.2890 & 0.3944  & 0.2630 \\
    0.9 & 0.6891 & 0.4395 & 0.5622 & 0.2611 & 0.3247 & 0.3029 & 0.5482  & 0.3419 \\
    \midrule
    Random 5 blocks & 0.7624 & 0.5072 & 0.6492 & 0.3119 & 0.3797 & 0.3042 & 0.5842  & 0.3605 \\
    All blocks & 0.2360 & 0.2736 & 0.2581 & 0.0495 & 0.1522 & 0.2928 & 0.2448  & 0.2233 \\
    Ours    & \textbf{0.7804} & \textbf{0.5482} & \textbf{0.6980} & \textbf{0.3280} & \textbf{0.3900} & \textbf{0.3054} & 0.5860  & \textbf{0.3691} \\
    \bottomrule
    \end{tabular}%
    }
    \label{tab:appendix_enhance_layers_selection}

\end{table}

\subsection{More Ablation Studies on Scale Schemes}

We compare various enhancement scale schemes $\lambda(l)$, including uniform ($U(1.2,1.8)$, $U(1.8,1.2)$), exponential ($Exp(1.6,0.95)$), and fixed ($1.5$) scales. As shown in Tab.~\ref{tab:scale_schemes}, all schemes outperform the baseline, reflecting the robustness of our block selection. The fixed scale offers simple and balanced results, while varying schemes exhibit specific strengths: increasing scales benefit spatial aspects, and decaying scales (e.g., $U(1.8,1.2)$) enhance color and shape. This suggests that both the magnitude and distribution of enhancement influence compositional generation.

\begin{table}[htbp]
    \centering
    \caption{Comparison of different scale schemes across blocks on T2I-CompBench++.}
    \resizebox{\linewidth}{!}{%
        \setlength{\tabcolsep}{4pt}
        \begin{tabular}{@{}ccccccccc@{}}
        \toprule
        \multicolumn{1}{c}{\multirow{2}{*}{\textbf{Methods}}} &
        \multicolumn{3}{c}{\textbf{Attribute Binding}} &
        \multicolumn{3}{c}{\textbf{Object Relationship}} &
        \multicolumn{1}{c}{\multirow{2}{*}{\textbf{Amount}}} &
        \multicolumn{1}{c}{\multirow{2}{*}{\textbf{Complex}}} \\ 
        \cmidrule(lr){2-4} \cmidrule(l){5-7}
        \multicolumn{1}{c}{} &
        \multicolumn{1}{c}{Color} &
        \multicolumn{1}{c}{Shape} &
        \multicolumn{1}{c}{Texture} &
        \multicolumn{1}{c}{2D Spatial} &
        \multicolumn{1}{c}{3D Spatial} &
        \multicolumn{1}{c}{Non-Spatial} &
        \multicolumn{1}{c}{} &
        \multicolumn{1}{c}{} \\
        \midrule
        FLUX     & 0.7322 & 0.4908 & 0.6490 & 0.2935 & 0.3739 & 0.3044 & \textbf{0.5877}  & 0.3597 \\
        U(1.2,1.8)    & 0.7795 & 0.5438 & 0.6827 & \textbf{0.3311} & 0.3865 & 0.3051 & 0.5845  & 0.3685 \\
        U(1.8,1.2)    & \textbf{0.8035} & \textbf{0.5551} & 0.6940 & 0.3203 & 0.3893 & 0.3051 & 0.5654  & \textbf{0.3722} \\
        Exp(1.6,0.95) & 0.7783 & 0.5507 & \textbf{0.7049} & 0.3296 & \textbf{0.3947} & \textbf{0.3060} & 0.5763  & 0.3710 \\
        1.5 (Fixed)    & 0.7804 & 0.5482 & \textbf{0.6980} & 0.3280 & 0.3900 & 0.3054 & 0.5860  & 0.3691 \\
        \bottomrule
        \end{tabular}%
    }
    \label{tab:scale_schemes}
\end{table}

\subsection{Generalization to Unseen Attributes}
\label{appendix:generalization_unseen_attributes}

To verify that the identified tailoring blocks generalize beyond the three representative attributes (color, shape, spatial) used during probing analysis, we construct two additional out-of-domain test sets: \textbf{Lighting} and \textbf{Art Style}, each containing $300$ prompts. These attributes were not seen during the block selection process.

As shown in Tab.~\ref{tab:appendix_generalization}, our method consistently improves over the baseline on both attributes, with a $+9.57\%$ gain on Lighting and $+10.14\%$ gain on Art Style. This confirms that the selected tailoring blocks capture general text-semantic processing regions rather than attribute-specific patterns.

\begin{table}[htbp]
    \centering
    \caption{Generalization to unseen attributes. Results are evaluated on FLUX with 300 prompts per attribute using VQA-based accuracy.}
    \label{tab:appendix_generalization}
    \setlength{\tabcolsep}{12pt}
    \begin{tabular}{lcc}
        \toprule
        Method & Lighting & Art Style \\
        \midrule
        Baseline & 0.4867 & 0.6567 \\
        Ours     & \textbf{0.5333} & \textbf{0.7233} \\
        \bottomrule
    \end{tabular}
\end{table}

\clearpage
\newpage
\section{More Qualitative Results}
\label{appendix:more_qualitative_results}

\subsection{More SD3.5 Results}

\begin{figure}[htbp]
    \vspace{-10pt}
    \centering
    \includegraphics[page=3,width=0.8\textwidth]{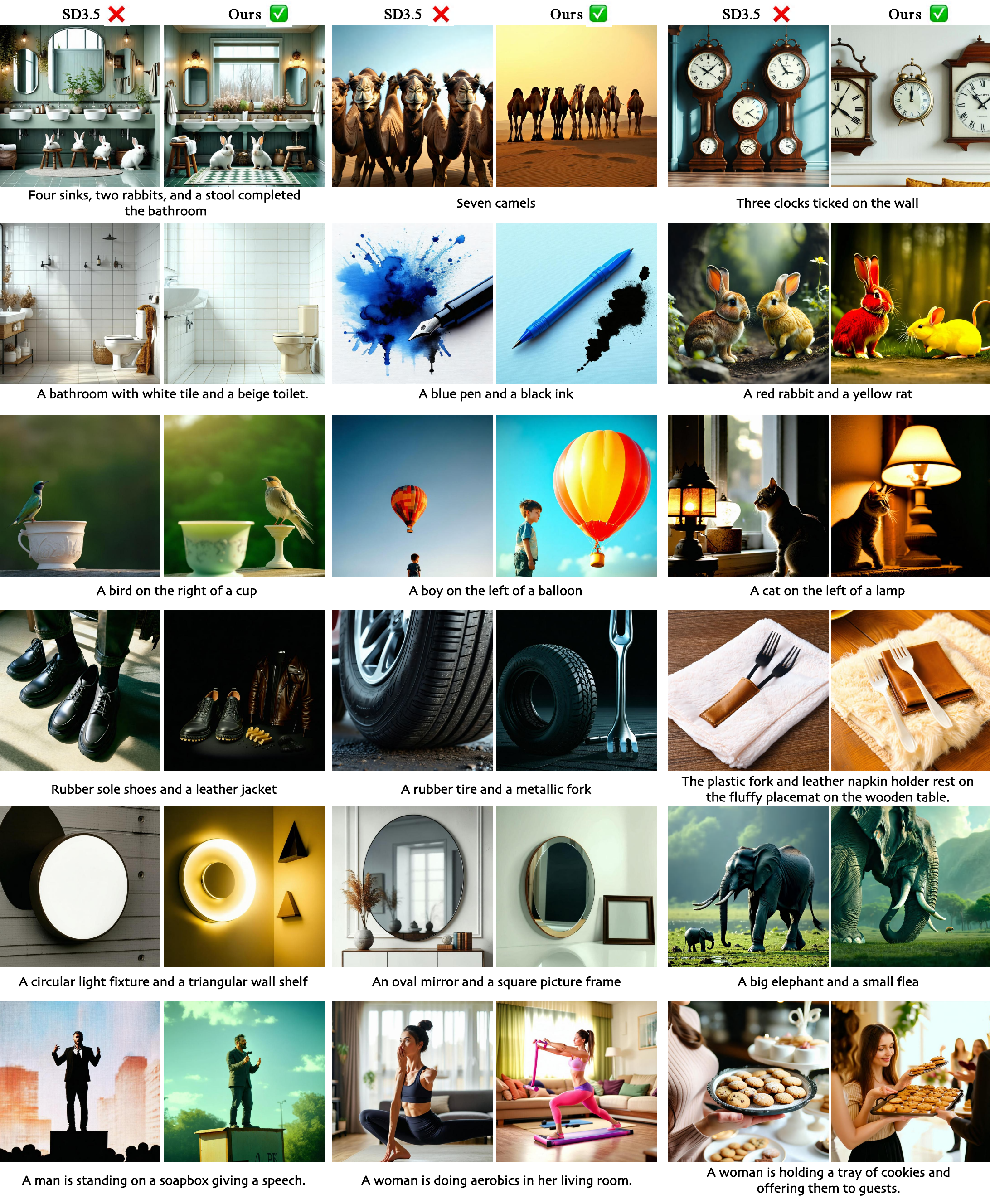}
    \caption{More qualitative results of SD3.5 and our method, covering aspects such as amount, color, spatial arrangement, texture, shape, and non-CLIP attributes. Our method consistently demonstrates better text alignment.}
    \label{fig:appendix_sd3_results}
    \vspace{-10pt}
\end{figure}

\clearpage
\newpage
\subsection{More FLUX Results}

\begin{figure}[htbp]
    \centering
    \includegraphics[page=4,width=0.8\textwidth]{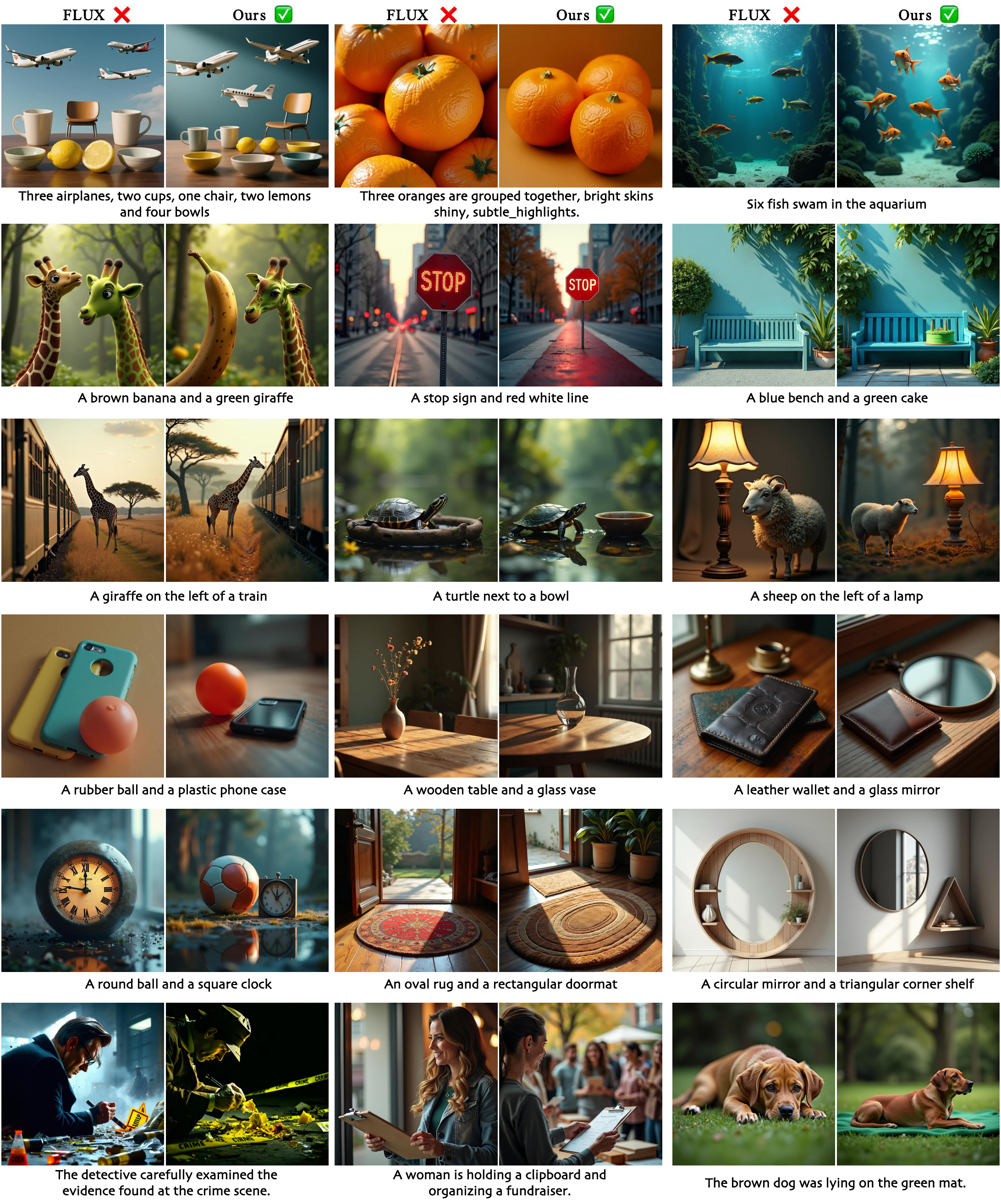}
    \caption{More qualitative results of FLUX and our method. Our approach achieves better text-image alignment and, in some cases, improved aesthetics over the baseline.}
    \label{fig:appendix_flux_results}
    \vspace{-10pt}
\end{figure}

\clearpage
\newpage
\subsection{More Qwen Image Results}

\begin{figure}[htbp]
    \centering
    \includegraphics[page=5,width=0.8\textwidth]{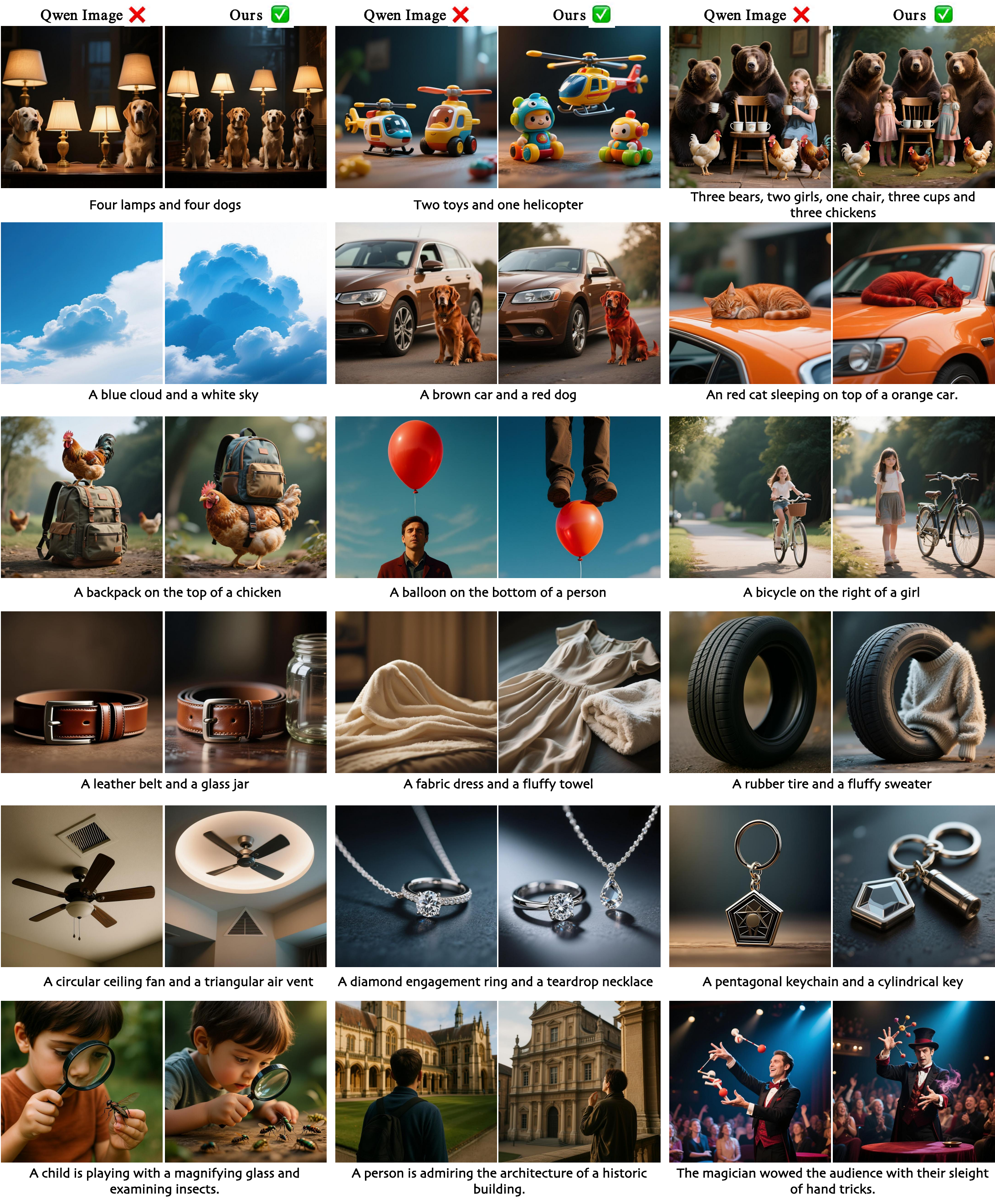}
    \caption{More qualitative results of Qwen Image and our method. Although the baseline fails on "A blue cloud and a white sky", our method succeeds.}
    \label{fig:appendix_qwen_results}
    \vspace{-10pt}
\end{figure}

\clearpage
\newpage
\subsection{More Editing Results}
\begin{figure}[htbp]
    \centering
    \includegraphics[page=6,width=0.95\textwidth]{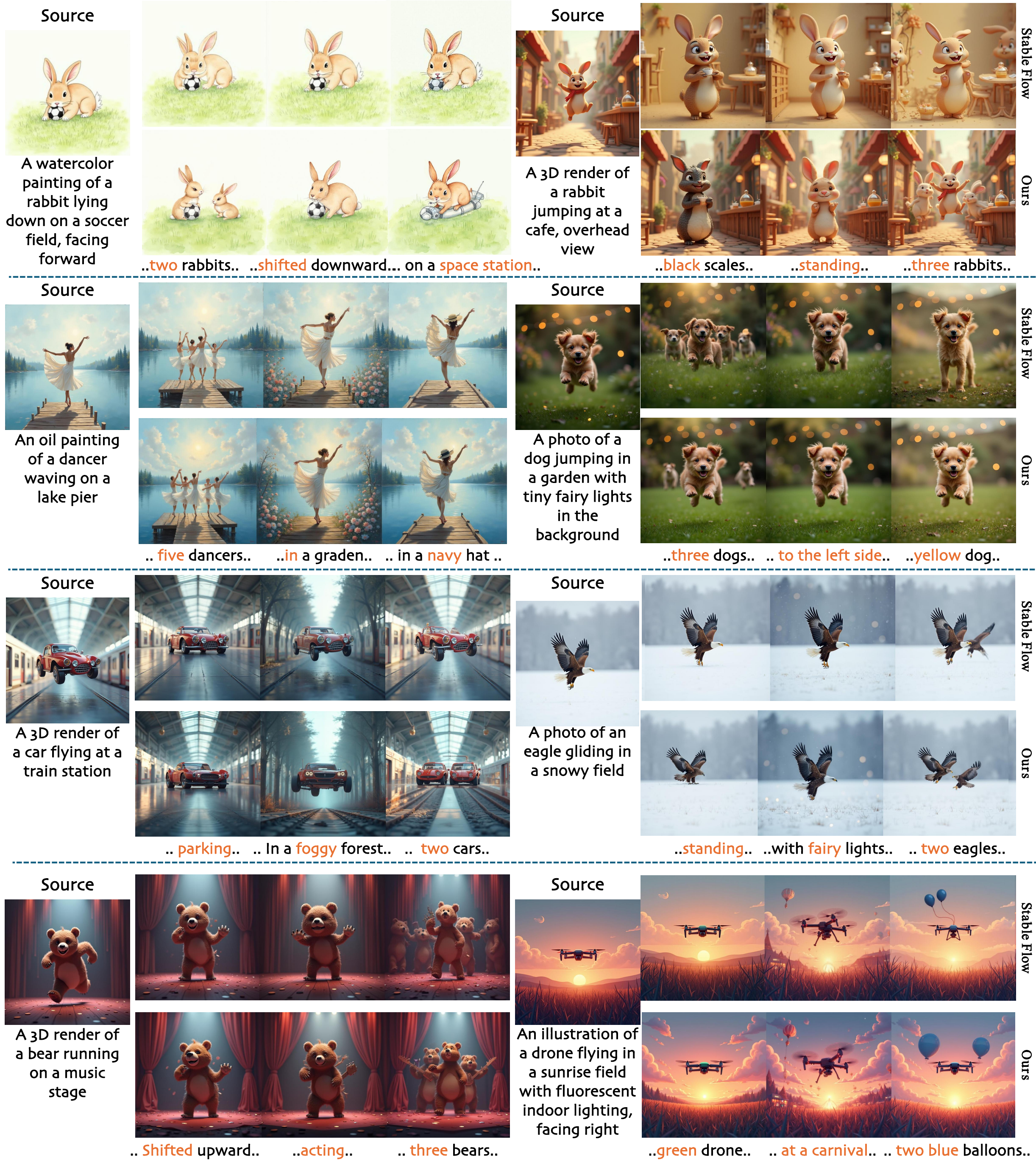}
    \caption{More editing examples on FLUX with our method and StableFlow. Our approach particularly surpasses StableFlow in quantity while maintaining high fidelity and strong text-image alignment.}
    \label{fig:appendix_editing_examples}
\end{figure}

\clearpage
\newpage
\subsection{Failure Cases}

\clearpage
\newpage
\subsection{All-block Showcases of Probing Analysis}

\begin{figure}[hbtp]
    \centering
    \label{fig:appendix_allayer_showcase}

    \begin{subfigure}{0.9\textwidth}
        \centering
        \includegraphics[page=2,width=0.9\textwidth]{figures/appendix_allayer_comp.pdf}
        \caption{\textit{removing} all-block showcase.}
        \label{fig:appendix_sd35_remove_showcase}
    \end{subfigure}

    \begin{subfigure}{0.9\textwidth}
        \centering
        \includegraphics[page=3,width=0.9\textwidth]{figures/appendix_allayer_comp.pdf}
        \caption{\textit{disabling} all-block showcase.}
        \label{fig:appendix_sd35_disable_showcase}
    \end{subfigure}

    \begin{subfigure}{0.9\textwidth}
        \centering
        \includegraphics[page=1,width=0.9\textwidth]{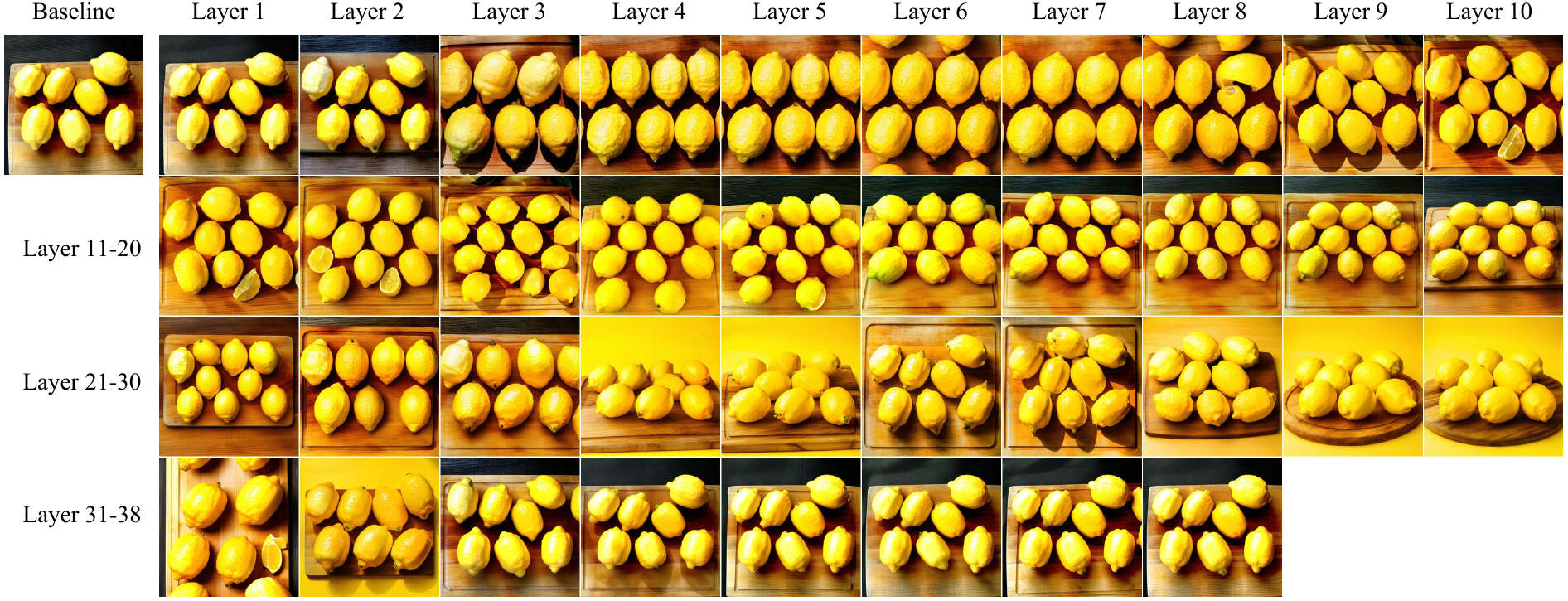}
        \caption{\textit{enhancing} all-block showcase.}
        \label{fig:appendix_sd35_enhance_showcase}
    \end{subfigure}

    \caption{All-block showcases of probing analysis on Stable Diffusion 3.5-large. From top to bottom, the prompts are: (a).``A white rabbit is hopping to the right of a brown basket.'',(b).``A red car with golden rims speeds along a coastal road.'', (c).``Seven lemons are arranged on a wooden cutting board, skins textured, color bright.''}
\end{figure}

\begin{figure}[h]
    \centering
    \label{fig:appendix_allayer_showcase}

    \begin{subfigure}{0.9\textwidth}
        \centering
        \includegraphics[page=5,width=0.8\textwidth]{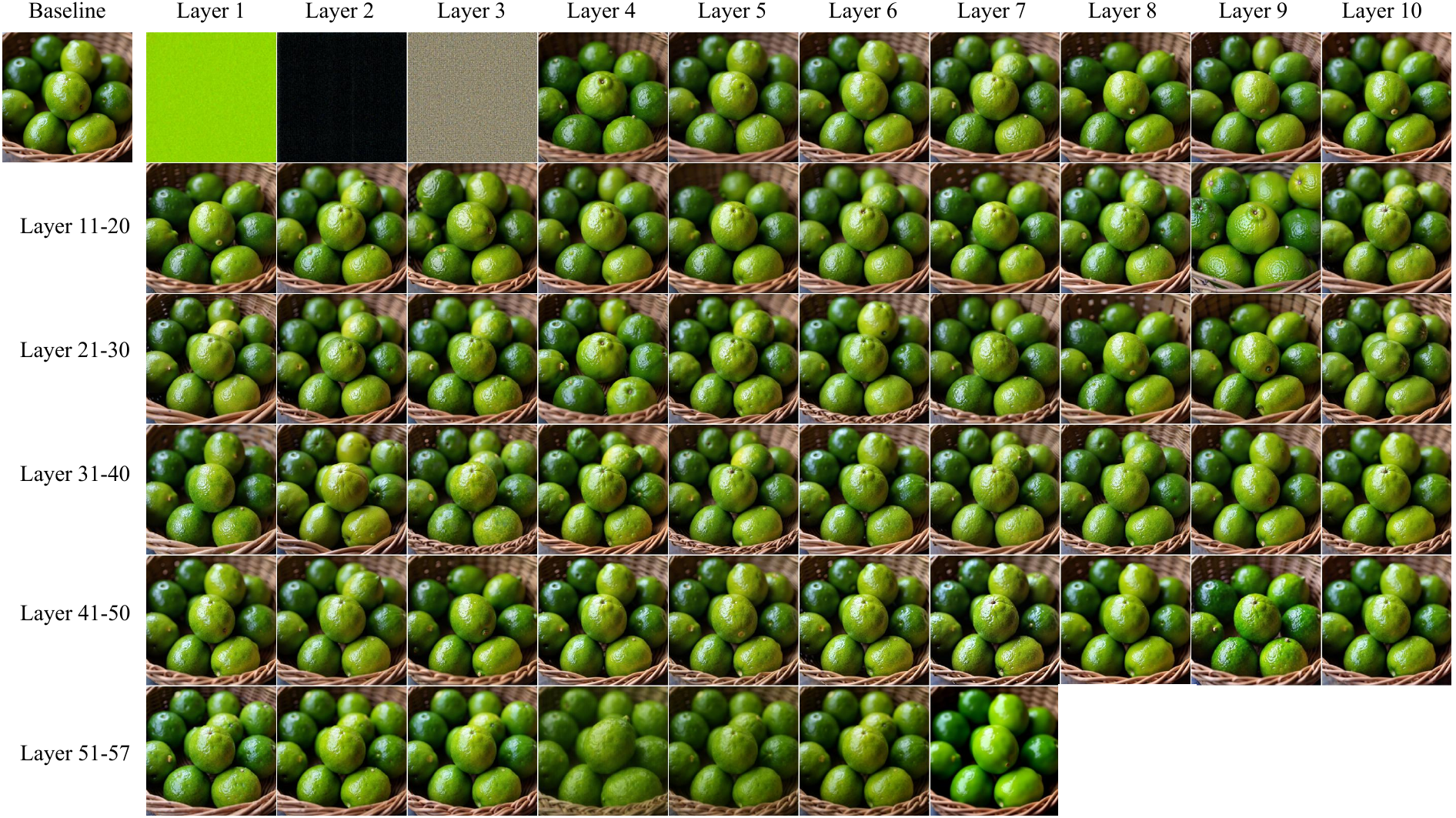}
        \caption{\textit{removing} all-block showcase.}
        \label{fig:appendix_flux_remove_showcase}
    \end{subfigure}

    \begin{subfigure}{0.9\textwidth}
        \centering
        \includegraphics[page=6,width=0.8\textwidth]{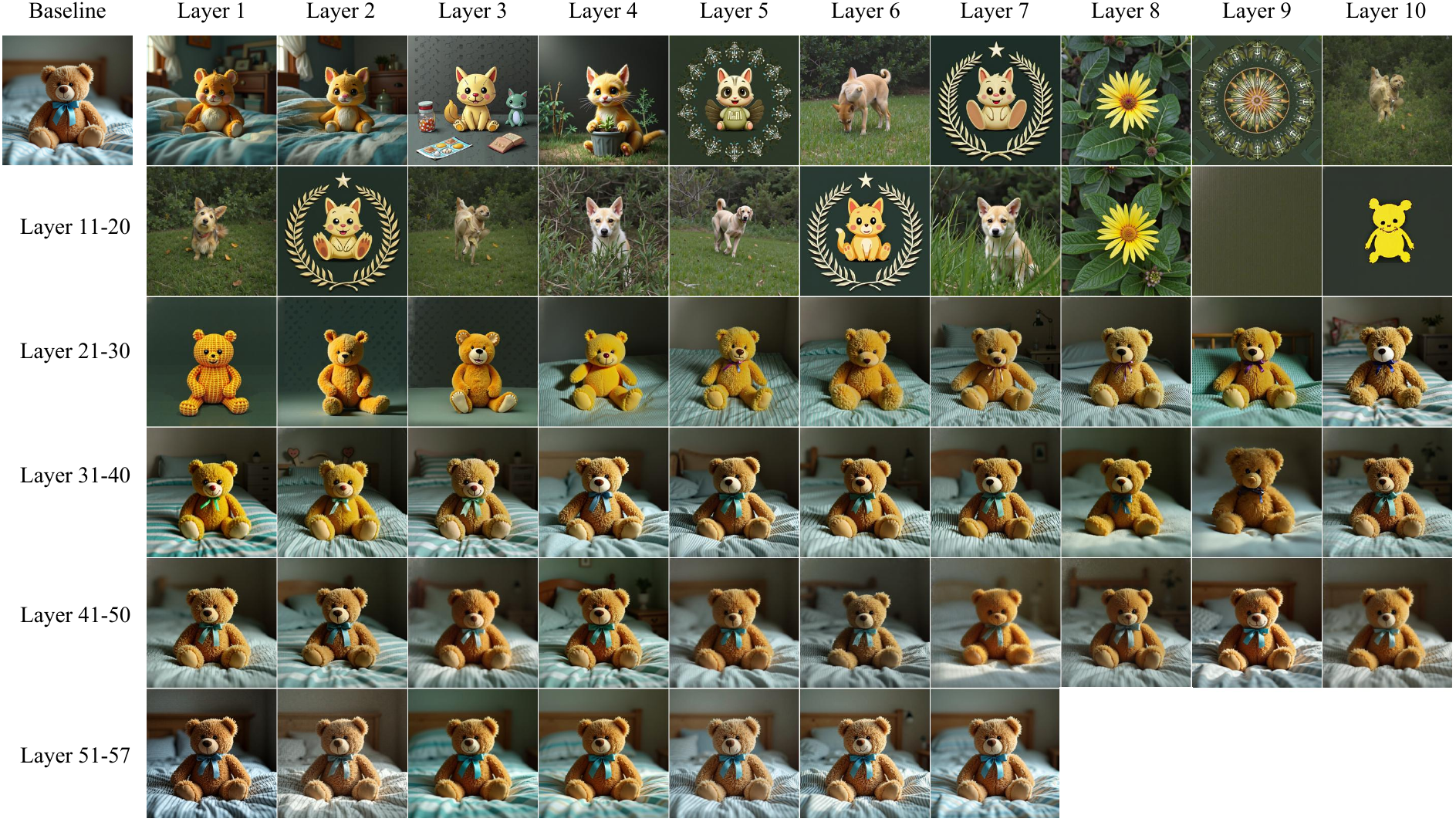}
        \caption{\textit{disabling} all-block showcase.}
        \label{fig:appendix_flux_disable_showcase}
    \end{subfigure}

    \begin{subfigure}{0.9\textwidth}
        \centering
        \includegraphics[page=4,width=0.8\textwidth]{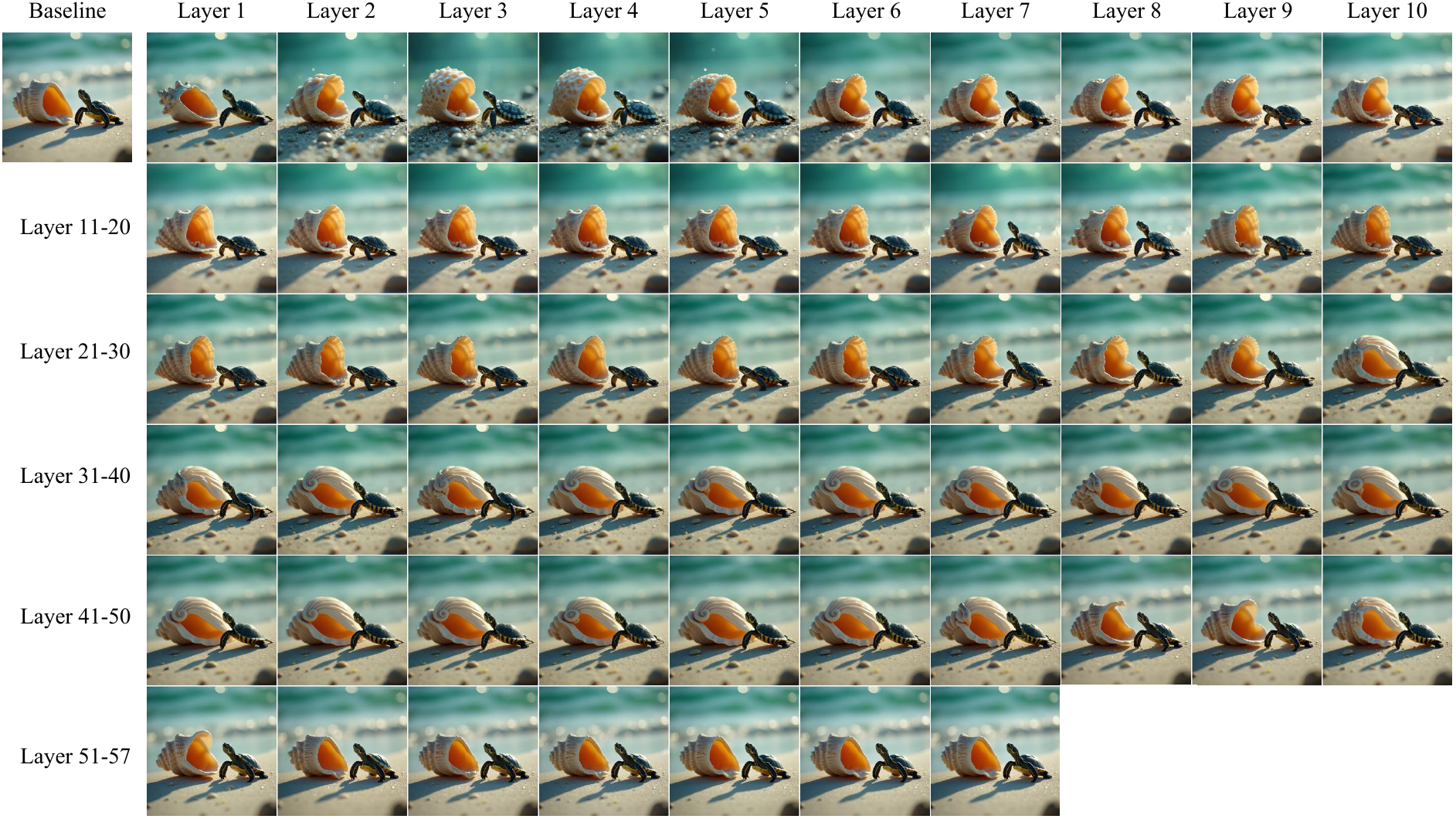}
        \caption{\textit{enhancing} all-block showcase.}
        \label{fig:appendix_flux_enhance_showcase}
    \end{subfigure}
    
    \caption{All-block showcases of probing analysis on FLUX.1-Dev. From top to bottom, the prompts are: (a).``Eight green limes rest in a basket, shiny skins and small droplets of water visible.'',(b).``A brown teddy bear with a blue ribbon sits on a child's bed with striped sheets.'', (c).``A small turtle moves to the right of a seashell.''}
\end{figure}

\begin{figure}[h]
    \centering
    \label{fig:appendix_allayer_showcase}

    \begin{subfigure}{0.9\textwidth}
        \centering
        \includegraphics[page=8,width=0.8\textwidth]{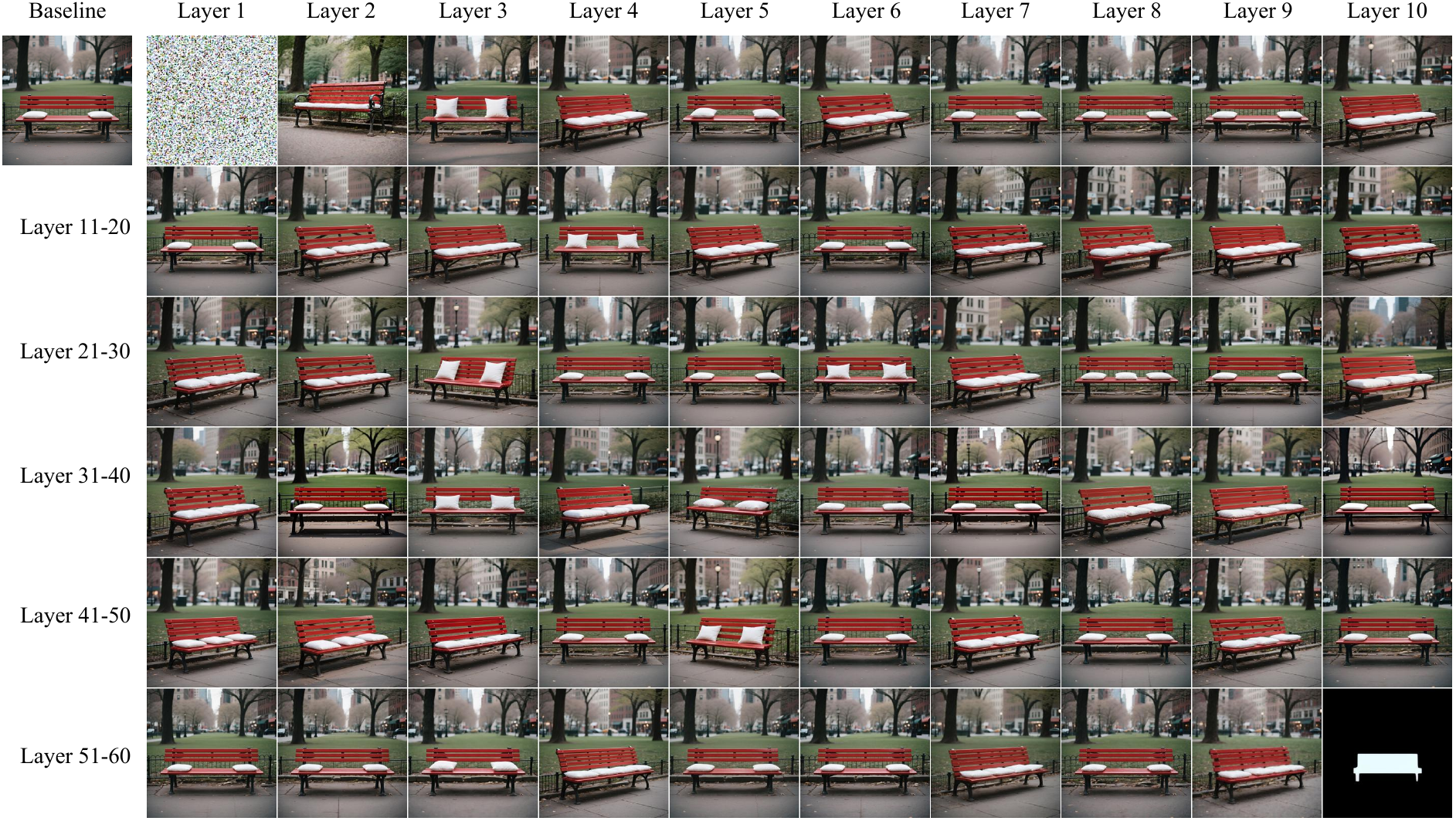}
        \caption{\textit{removing} all-block showcase.}
        \label{fig:appendix_qwen_remove_showcase}
    \end{subfigure}

    \begin{subfigure}{0.9\textwidth}
        \centering
        \includegraphics[page=9,width=0.8\textwidth]{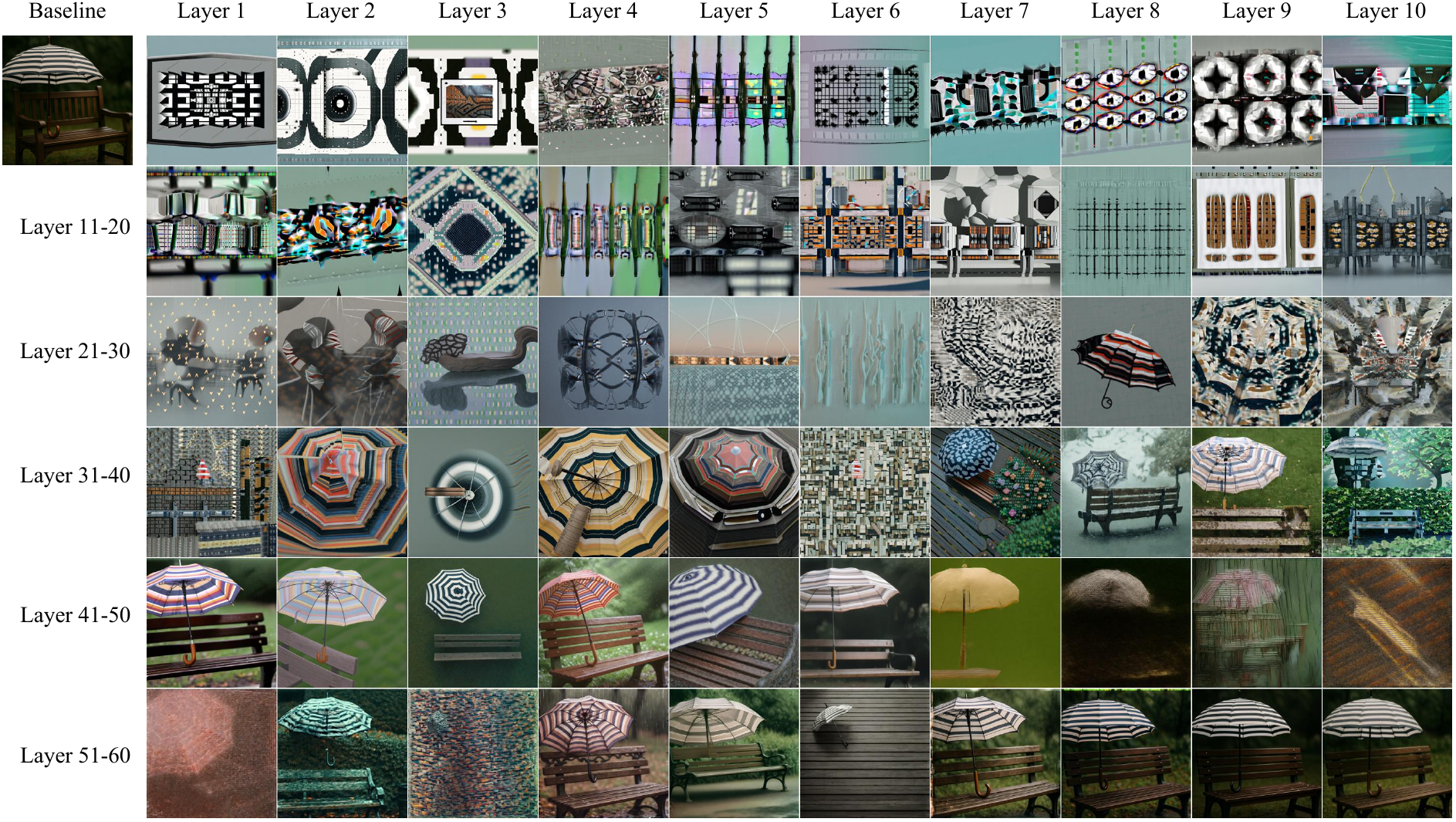}
        \caption{\textit{disabling} all-block showcase.}
        \label{fig:appendix_qwen_disable_showcase}
    \end{subfigure}

    \begin{subfigure}{0.9\textwidth}
        \centering
        \includegraphics[page=7,width=0.8\textwidth]{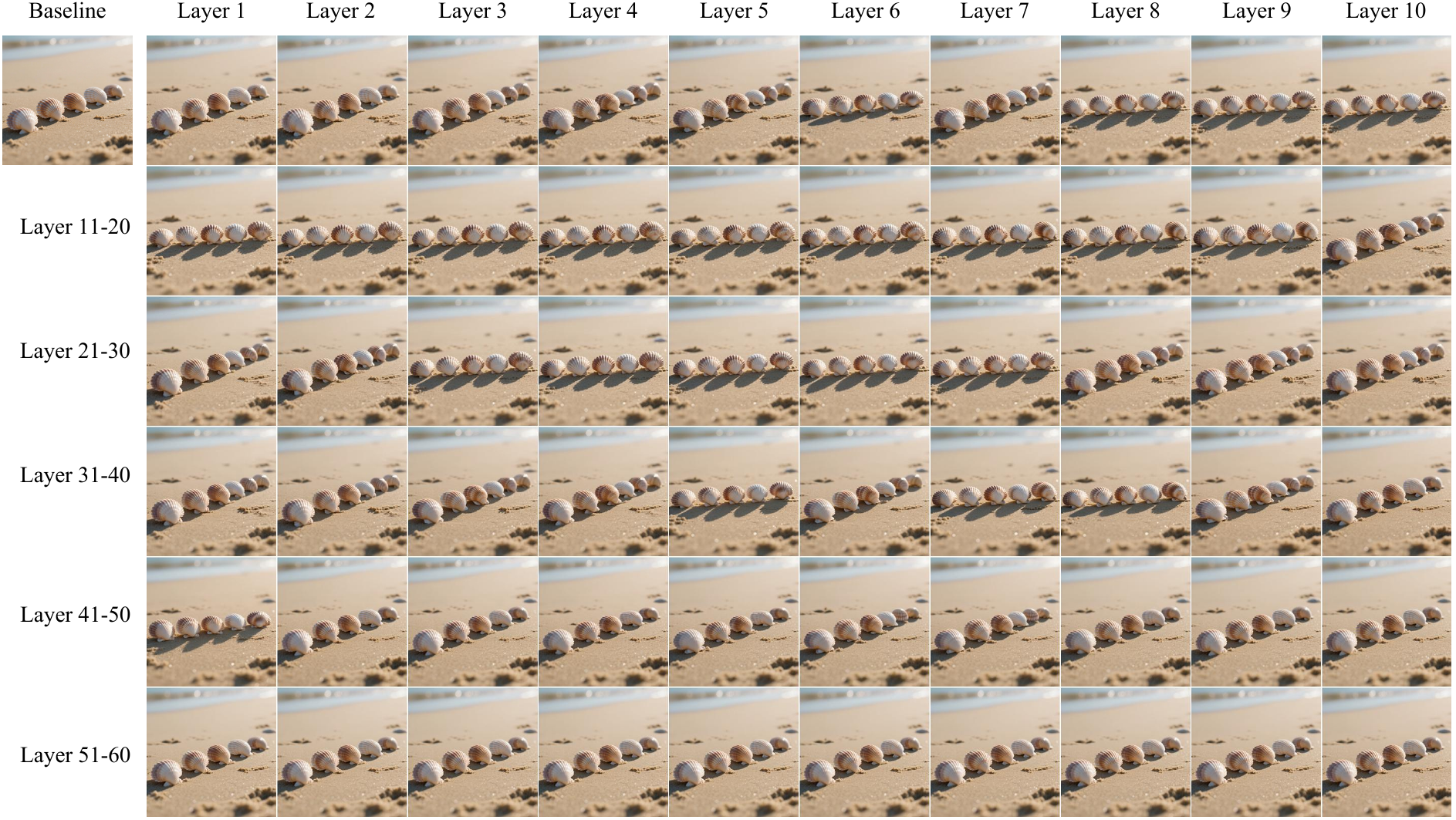}
        \caption{\textit{enhancing} all-block showcase.}
        \label{fig:appendix_qwen_enhance_showcase}
    \end{subfigure}

    \caption{All-block showcases of probing analysis on Qwen Image. From top to bottom, the prompts are: (a).``A red bench with white cushions stands in a quiet city park.'',(b).``A striped umbrella is positioned at the top-left of a wooden bench.'', (c).``Five seashells lie in a row on sand, each detailed with ridges and soft reflections.''}
\end{figure}

\clearpage
\newpage
\section{Human Evaluation Details}
\label{appendix:human_evaluation_details}

\begin{figure}[t]
    \centering
    \begin{subfigure}{0.48\textwidth}
        \centering
    \includegraphics[page=1,width=0.8\textwidth]{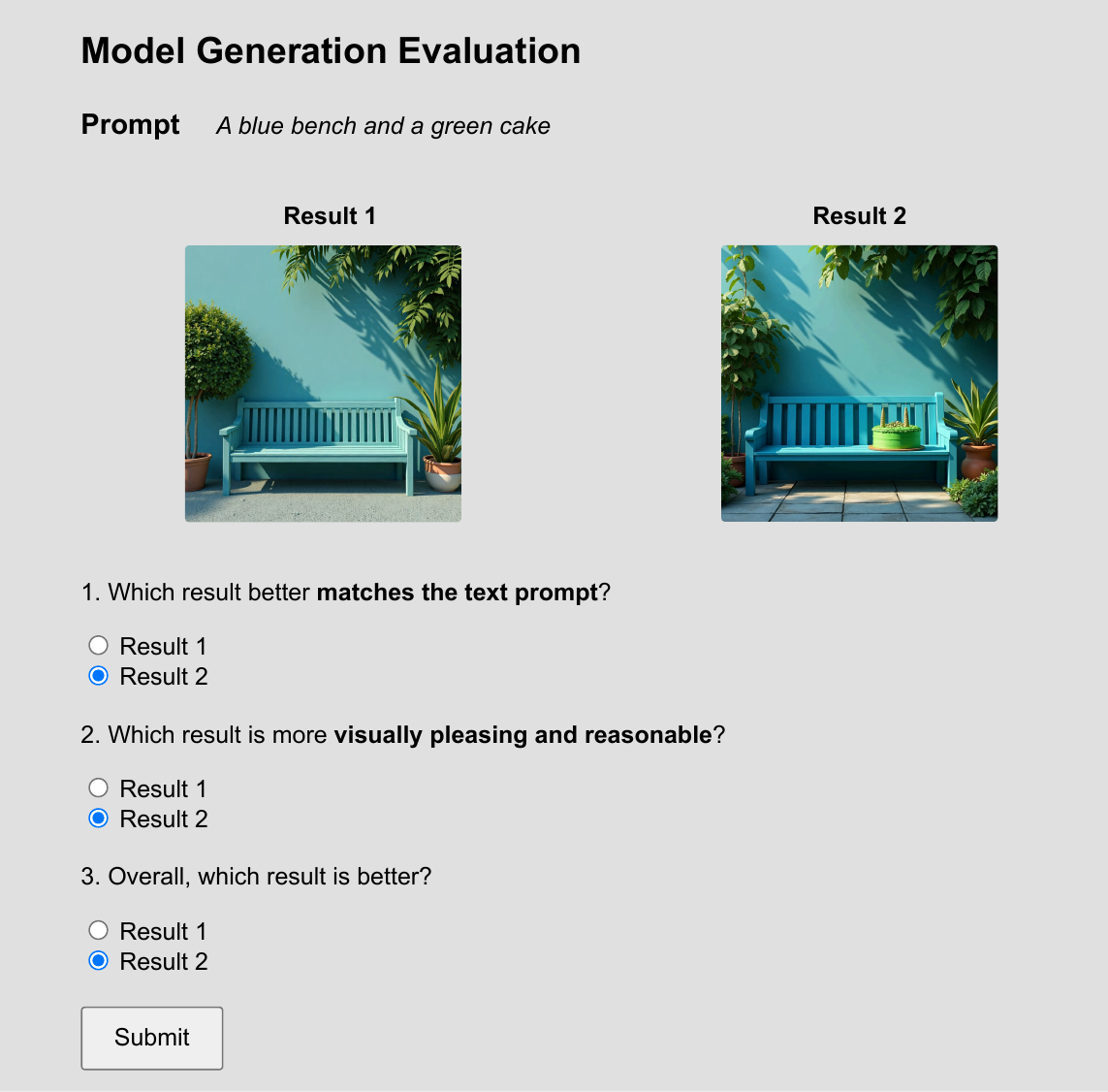}
    \caption{Interface for generation evaluation.}
    \label{fig:appendix_humaneval_generation}
    \end{subfigure}
    \hfill
    \begin{subfigure}{0.50\textwidth}
        \centering
        \includegraphics[page=2,width=0.8\textwidth]{figures/appendix_human_eval_interface.pdf}
        \caption{Interface for editing evaluation}
        \label{fig:appendix_humaneval_editing}
    \end{subfigure}
    \caption{
        \textbf{Interfaces for human evaluation.}
        Participants are asked three questions regarding alignment, preservation, and overall quality.}
    \label{fig:appendix_humaneval_interface}
\end{figure}

We conduct human evaluations for both generation and editing tasks.
For text-to-image generation, we sample $100$ prompts from T2I-Compbench++ and generate results using FLUX.1-Dev and our method. Participants are provided with the prompt and the two generated images in random order, and answer the questions in Fig.~\ref{fig:appendix_humaneval_generation}. The human preference score is defined as
\[
\text{Human Preference} = 1/4*\text{Alignment} + 1/4*\text{Aesthetic} + 1/2*\text{Overall}.
\]
For image editing, we randomly select $100$ samples from the editing dataset and apply our method and the baseline~\cite{stableflow_avrahami2025} with identical prompts. Three participants are recruited and, given the original image, the editing prompt, and two edited results in random order, they answer the questions in Fig.~\ref{fig:appendix_humaneval_editing}. The final preference score is computed as 
{\small
\[
\text{Human Preference} = 1/4*\text{Alignment} + 1/4*\text{Preservation} + 1/2*\text{Overall}.
\]
}
The corresponding quantitative results are summarized in Tab.~\ref{tab:human_evaluation_generation} and Tab.~\ref{tab:human_evaluation_editing}.

\begin{table}[htbp]
    \vspace{-20pt}
    \scriptsize
    \centering
    \caption{Human evaluation results for generation.}
    \label{tab:human_evaluation_generation}
    \begin{tabular}{lcccc}
        \toprule
        Methods & Alignment & Aesthetic & Overall & Human Preference\\
        \midrule
        FLUX & 474 & 531 & 430 & 38.85\% \\
        + Ours & 726 & 669 & 770 & 61.15\% \\
        \bottomrule
    \end{tabular}
    \vspace{-20pt}
\end{table}

\begin{table}[htbp]
    \vspace{-20pt}
    \scriptsize
    \centering
    \caption{Human evaluation results for editing.}
    \label{tab:human_evaluation_editing}
    \begin{tabular}{lcccc}
        \toprule
        Methods & Alignment & Preservation& Overall & Human Preference\\
        \midrule
        StableFlow & 498 & 539 & 465 & 40.98\% \\
        Our Method & 702 & 661 & 735 & 59.02\% \\
        \bottomrule
    \end{tabular}
    \vspace{-10pt}
\end{table}

\end{document}